\documentclass{article}

\usepackage{microtype}
\usepackage{graphicx}
\usepackage{subcaption}
\usepackage{booktabs} 

\usepackage{tikz}
\usetikzlibrary{positioning,arrows.meta,fit}
\usetikzlibrary{calc}
\usetikzlibrary{intersections,positioning}
\usepackage{pgfplots}
\pgfplotsset{compat=1.18} 

\usepackage{arydshln}
\usepackage{diagbox}
\usepackage{makecell}

\usepackage{hyperref}

\usepackage{dblfloatfix}

\usepackage[accepted]{icml2026}

\usepackage{amsmath}
\usepackage{amssymb}
\usepackage{mathtools}
\usepackage{amsthm}

\usepackage[capitalize,noabbrev]{cleveref}

\theoremstyle{plain}

\theoremstyle{definition}

\theoremstyle{remark}

\usepackage{arydshln}
\usepackage{tcolorbox}
\tcbuselibrary{breakable}
\usepackage{enumitem}

\usepackage[textsize=tiny]{todonotes}

\icmltitlerunning{Unconventional Hardware Neural Architecture Search}

\begin{document}

\twocolumn[
  \icmltitle{LLM-Guided Neural Architecture Search\\ for Robust Co-Design of Physical Neural Networks}



  \icmlsetsymbol{equal}{*}

  \begin{icmlauthorlist}
    \icmlauthor{Tyler King}{yyy,pton}
    \icmlauthor{Timothee Leleu}{yyy,comp}
  \end{icmlauthorlist}

  \icmlaffiliation{yyy}{NTT Research}
  \icmlaffiliation{pton}{Princeton University}
  \icmlaffiliation{comp}{Stanford University}

  \icmlcorrespondingauthor{Tyler King}{tyler.king@ntt-research.com, tk9138@princeton.edu}

  \icmlkeywords{Neural Architecture Search, Unconventional Computing, Automated Machine Learning, Optical Computing}

  \vskip 0.3in
]



\makeatletter
\def\ICML@appearing{}
\makeatother
\printAffiliationsAndNotice{}  

\begin{abstract}
Deploying neural networks on unconventional hardware demands architectures that co-optimize task accuracy and platform-specific constraints such as energy cost, physical non-idealities, and numerical precision. Existing neural architecture search (NAS) methods are typically tailored to a single hardware family, limiting cross-platform comparison and generalization. We introduce Unconventional Hardware Neural Architecture Search (UH-NAS), a hardware-agnostic, LLM-guided NAS framework that integrates language models as evolutionary operators to co-optimize accuracy and inference energy. By exposing hardware as a swappable backend with per-platform energy models, physical constraints, and non-ideality simulators, UH-NAS enables fair system-level comparisons across various backends without modifying the search algorithm. Tested on optical MZI hardware, UH-NAS discovers more diverse, robust architectures than conventional baselines while outperforming existing LLM-to-NAS approaches. Additional ablations on architecture robustness under non-idealities and the role of system prompts highlight the importance of architecture–hardware co-design for emerging computing platforms.
\end{abstract}

\section{Introduction}

\begin{figure}[t]
  \centering
\includegraphics[width=\columnwidth]{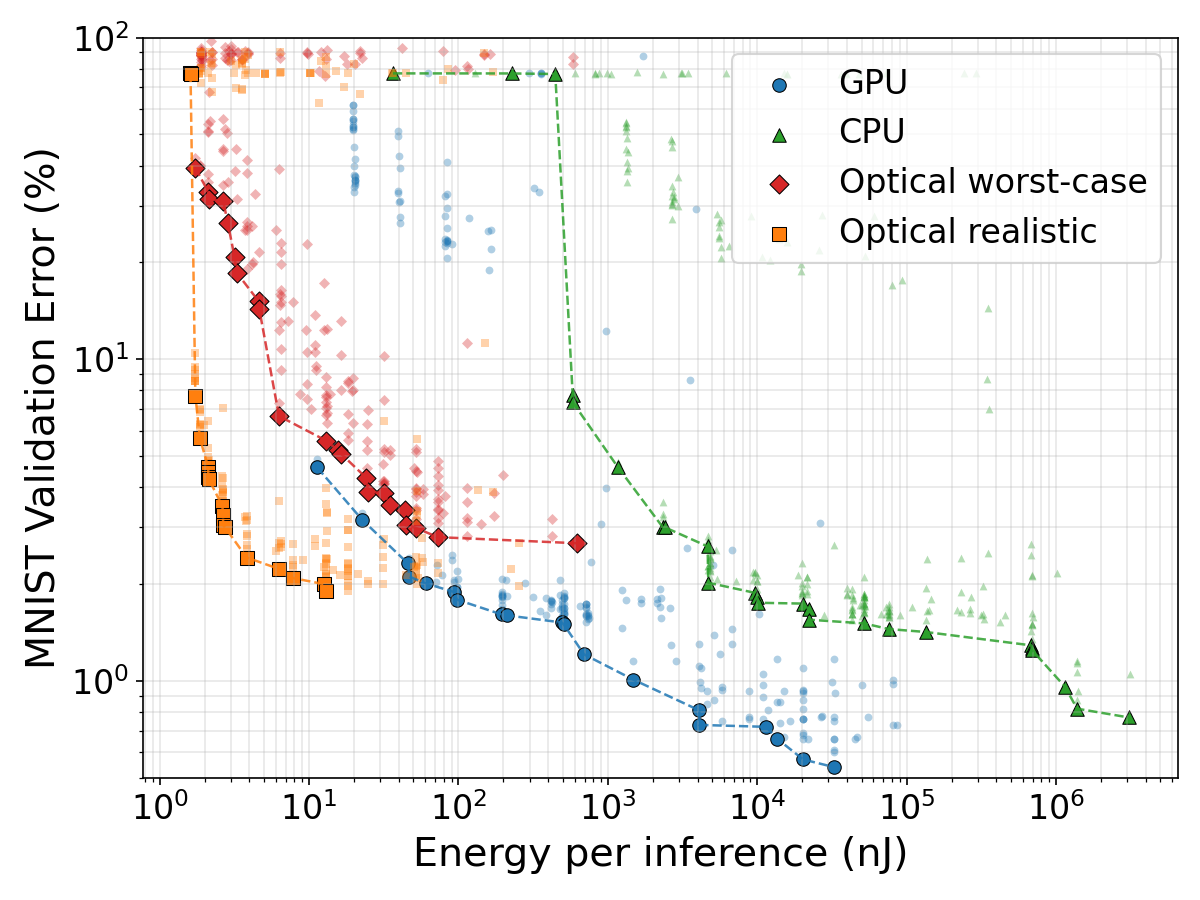}
    \caption{Pareto fronts of MNIST validation error (\%) versus energy per inference (nJ) obtained with UH-NAS across multiple hardware backends, including GPU (Blackwell B200, FP16), CPU (Xeon 8380, FP32), and MZI mesh–based optical systems under both realistic and worst-case non-idealities. Optical hardware operates at substantially lower energy but with higher error rates due to device non-idealities; UH-NAS identifies robust architectures that narrow this gap.
    }
  \label{fig:architecture}
\end{figure}

Neural architecture search (NAS) has become a standard framework for automating neural-network design, and a central lesson from hardware-aware NAS is that the best architecture depends on the target deployment platform rather than accuracy alone. Classical NAS work formalized the problem in terms of search space, search strategy, and performance estimation, while platform-aware methods showed that latency, energy, and device characteristics can change the optimal accuracy--cost trade-off \cite{elsken2019neural, tan2019mnasnet, li2021hw, li2021flash, jiang2020device, yuan2021nas4rram, zhou2021analognets, negi2022nax, bessalah2025analognas}. In other words, hardware does not merely rescale the score of a model, but may also change which architectural motifs are preferred.

This dependence on hardware is even stronger on unconventional hardware. In analog in-memory computing, non-ideal crossbars, and photonic tensor cores, performance is shaped not only by nominal throughput, but also by device non-idealities such as drift, noise, mismatch, crosstalk, and limited precision \cite{benmeziane2023analognas, yan2021uncertainty, bhattacharjee2023xplorenas, wang2022quantumnas, gu2022adept}. Photonic computing provides a particularly compelling example: optical neural hardware promises high-throughput and energy-efficient linear operations, yet its behavior can be highly sensitive to phase errors, calibration, and physical design choices \cite{Shen2017nanophotonics, hamerly2019large, shastri2021photonics}. Existing search methods for these settings are typically tailored to a single hardware family and rely on hand-designed objectives, search rules, or robustness models that may break down under slight hardware changes.

At the same time, recent work has shown that large language models (LLMs) can act as flexible search operators for architecture design, serving as mutation, reflection, and reasoning modules in evolutionary or quality-diversity NAS loops \cite{chen2023evoprompting, zheng2023can, nasir2024llmatic, ji2025rznas, zhu2025llmnas}. This suggests a different route to hardware-aware NAS: rather than designing a new NAS algorithm for every emerging hardware, one can use an LLM to integrate hardware-specific context and adapt its search behavior through simulator feedback. Motivated by this idea, we introduce Unconventional Hardware Neural Architecture Search (UH-NAS), a robust LLM-guided NAS for unconventional hardware. 

UH-NAS combines these two lines of work by integrating LLM-guided evolutionary search into a hardware-aware NAS framework for unconventional computing platforms. In contrast to prior NAS methods for unconventional hardware, which are often specialized to a single hardware paradigm with hand-designed objectives, UH-NAS models hardware through operation-level energy costs, physical constraints, and non-ideality models associated with standard neural-network operations, enabling the same search procedure to operate across heterogeneous platforms. Search is further guided by LLM-generated system prompts and a hardware-specific knowledge base updated after each NAS generation. Furthermore, unlike traditional NAS approaches that typically ignore hardware non-idealities and often rely on zero-cost (ZC) proxies (which we observe to break down under hardware non-idealities), we score every candidate via full noisy training, guided by the aforementioned hardware-specific design heuristics to help guide the search.

Our main results are as follows. First, the proposed framework enables consistent system-level comparison across heterogeneous hardware platforms (CPU, GPU, and MZI-based optical systems) by optimizing architectures separately for each hardware under its own constraints. Second, under strong hardware non-idealities, UH-NAS discovers more diverse architectures and achieves lower validation error than conventional NAS baselines. Third, architectures optimized under severe non-idealities differ qualitatively from those found in ideal settings, indicating that robustness cannot be extrapolated from ideal training. Finally, the method uncovers favorable accuracy–energy trade-offs on unconventional hardware such as optical systems, highlighting the importance of architecture–hardware co-design for revealing system-level advantages of emerging platforms. Analyzing such architectures yields physically interpretable design principles for unconventional hardware platforms, bridging the gap between NAS-discovered architectures and physical hardware understanding.



\section{Related Work}

\subsection{NAS for Unconventional Computing}

As unconventional computing platforms improve in both energy efficiency and accuracy, the need for machine-learning architectures capable of fully realizing the advantages of these emerging hardware systems continues to grow. Conventional NAS methods have been applied across a wide range of unconventional hardware paradigms to automate the discovery of architectures that fully exploit platform-specific strengths, notably in analog in-memory computing \cite{bessalah2025analognas, benmeziane2023analognas, bhattacharjee2023xplorenas, Krestinskaya2024imcnas}, photonics \cite{gu2022adept, Shafiee2025luxnas}, quantum computing \cite{wang2022quantumnas, wu2023quantumdarts, Martyniuk2024quantumnassurvey}, and spiking/neuromorphic computing \cite{na2022autosnn, Yan2024efficientsnn}.

Despite the promising performance of NAS for unconventional computing, existing approaches are often hardware-specific and rely on hand-designed objectives and heuristics, limiting their ability to adapt to varying hardware characteristics and non-idealities without hardware-specific expertise. Recent LLM-assisted search and algorithm-discovery methods provide a promising direction for overcoming these limitations by incorporating hardware-specific knowledge directly into the search process.


\subsection{Evolutionary Algorithms and Algorithm Discovery}

LLM-driven evolution has recently emerged as a powerful paradigm for autonomous algorithm discovery, with FunSearch \cite{romera2024mathematical}, AlphaEvolve \cite{novikov2025alphaevolve}, and AlphaTensor \cite{fawzi2022discovering} producing breakthrough results on long-standing mathematical and algorithmic problems, while EoH \cite{liu2024evolution} and ReEvo \cite{ye2024reevo} extend the paradigm to combinatorial heuristic design through refinement loops. Building on this line of work, we adopt NSGA-II \cite{Deb2002nsgaii} as our evolutionary backbone. It natively handles the multi-objective nature of NAS, allowing UH-NAS to construct high-performing accuracy--energy Pareto fronts for cross-hardware comparisons while also preserving population diversity. 



\subsection{LLM-assisted NAS}

Recent advances in LLM-assisted algorithm discovery naturally extend to neural architecture search on conventional hardware platforms. These LLM-to-NAS approaches typically utilize LLMs to either predict components as part of evolutionary architectures \cite{chen2023evoprompting, yu2025gptnas} or prompt them to propose novel architectures \cite{zhu2025llmnas, ji2025rznas, nasir2024llmatic, zheng2023can}. To the best of our knowledge, however, these LLM-guided NAS approaches have primarily been studied on conventional digital electronics (e.g., CPU/GPU), with comparatively limited exploration in unconventional hardware.

To accelerate their search, many such NAS algorithms leverage ZC (zero cost) proxies for architecture feedback \cite{krishnakumar2022nasbenchsuitezero, zhu2025llmnas, ji2025rznas}, allowing for model evaluation at initialization. Despite this promise, we observe that ZC proxies break down under hardware non-idealities, which are particularly prevalent in unconventional computing schemes \cite{hamerly2019large, bhattacharjee2023xplorenas, bessalah2025analognas}. 

Critically, unconventional hardware breaks key assumptions underlying existing LLM-to-NAS methods: zero-cost proxies, widely adopted to accelerate NAS search \cite{krishnakumar2022nasbenchsuitezero, zhu2025llmnas, ji2025rznas}, exhibit negative correlations with validation accuracy under hardware non-idealities (Figure 5), hardware-specific constraints invalidate large portions of the search space (e.g. prohibitively expensive skip connections), and an architecture's clean performance is not predictive of its noisy performance \cite{hamerly2019large, bhattacharjee2023xplorenas, bessalah2025analognas}. Simply integrating a new energy model into existing frameworks is insufficient to address such issues as it would fail to incorporate constrained search spaces, non-ideality-aware evaluation or hardware-specific knowledge. UH-NAS addresses such issues through its hardware-agnostic backend abstraction, full noisy training of every candidate, and LLM-synthesized hardware-specific design heuristics.

\section{Method}

\subsection{Problem Formulation}

Our optimization goal is to find candidate architectures that generate a Pareto front of our proposed multi-objective optimization problem given a specific hardware type. For each hardware backend $i \in \{\mathrm{optical}, \mathrm{gpu}, \mathrm{cpu}\}$, UH-NAS evaluates a population of candidate architectures $\{x_{i,k}\}_{k=1}^{N_i}$, where $k$ indexes architectures searched for that backend. For notational simplicity, we write $x_i$ when referring to a generic candidate architecture under hardware backend $i$. More formally, for a candidate architecture $x_i$ evaluated on hardware backend $i$, we seek to compute $\min_{x_i} F_i(x_i)$, where
\begin{align}
F_i(x_i)
=
\bigl(
f_{i,1}(x_i),\;
f_{i,2}(x_i),\;
\dots
\bigr)
\end{align}
denotes a vector of hardware-dependent objective functions.
From this, we construct a Pareto front, defined as the set of architectures $x_i$ that are not strictly dominated in this objective space (i.e.\ for which no other candidate architecture achieves both higher accuracy and lower energy). More formally, the Pareto front $\mathcal{P}_i$ is defined as:

\begin{align}
\mathcal{P}_i =\large\{ & F_i(x_i)\;:\;x_i\in\mathcal{X}_i,\; \nexists\,x_i'\in\mathcal{X}_i\nonumber \\ 
& \text{ s.t. } F_i(x_i') \preceq F_i(x_i) \text{ and } F_i(x_i') \neq F_i(x_i) \large\}.
\end{align}






\paragraph{Proof-of-concept instantiation setup}

For the proof-of-concept considered in this work, we restrict the optimization problem to maximizing classification accuracy $f_{i,\text{acc}}(x_i)$ and minimizing inference energy $f_{i,\text{eff}}(x_i)$. More formally,
\begin{align}
F_i(x_i)
=
\bigl(
-f_{i,\mathrm{acc}}(x_i),\;
f_{i,\mathrm{eff}}(x_i)
\bigr),
\end{align}
\noindent where the inference energy $f_{i,\mathrm{eff}}(x_i)$ is approximated using platform-specific analytic cost models based on the structure of the model. 

We instantiate the proposed framework on MNIST classification, where a candidate $x_i$ is a PyTorch program defining a neural network. Given a candidate $x_i$, we train the model and evaluate the classification accuracy $f_{i,\mathrm{acc}}(x_i;\mathcal{B}_i)$ on a validation set. The training procedure is mildly hardware-aware: depending on the hardware regime $\mathcal{B}_i$ (consisting of hardware non-idealities, constraints, energy model), certain layers are replaced by quantization-aware variants (e.g., 8-bit optical models), providing a first-order approximation of hardware effects. We additionally integrate hardware constraints (e.g., expensive skip connections in optical computing) and hardware non-idealities (e.g. phase error, crosstalk in optical computing). A full list of hardware constraints and non-idealities is provided in Section \ref{sec:hardware_considerations}.


This implementation provides a minimal but functional realization of hardware-aware LLM-to-NAS, where architectures are jointly evaluated with respect to task loss and hardware-dependent cost.

\subsection{UH-NAS Implementation}



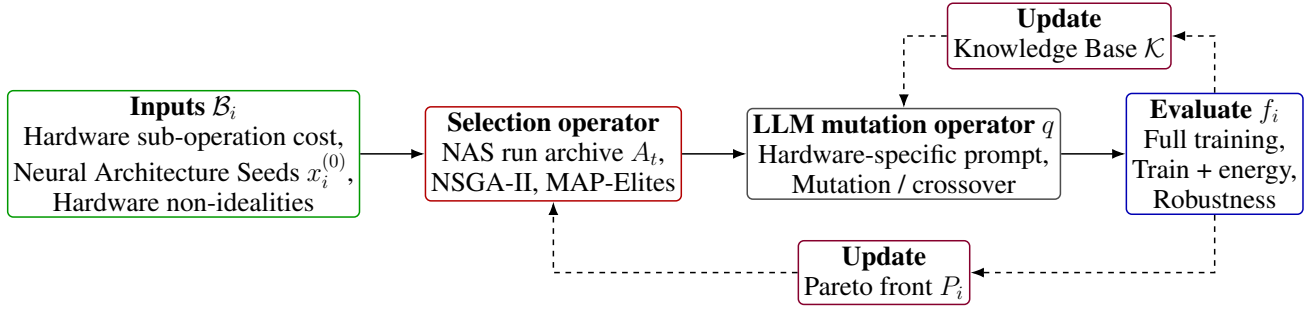
\begin{figure*}[t]
\centering
\resizebox{\textwidth}{!}{%
\begin{tikzpicture}[
    font=\Large,
    box/.style={
        draw,
        rounded corners=3pt,
        align=center,
        minimum width=3.2cm,
        minimum height=1.0cm,
        inner sep=3pt,
        thick
    },
    smallbox/.style={
        draw,
        rounded corners=3pt,
        align=center,
        minimum width=2.8cm,
        minimum height=0.9cm,
        inner sep=3pt,
        thick
    },
    inputbox/.style={box, draw=green!60!black},
    searchbox/.style={box, draw=red!70!black},
    llmbox/.style={box, draw=black!70},
    evalbox/.style={box, draw=blue!70!black},
    updatebox/.style={smallbox, draw=purple!70!black},
    arrow/.style={-{Latex[length=2.5mm]}, thick},
    feedback/.style={-{Latex[length=2.5mm]}, thick, dashed}
]
\node[inputbox] (inputs) {
    \textbf{Inputs $\mathcal{B}_i$}\\
    Hardware sub-operation cost,\\
    Neural Architecture Seeds $x^{(0)}_i$,\\
    Hardware non-idealities
};
\node[searchbox, right=1.2cm of inputs] (controller) {
    \textbf{Selection operator}\\
    NAS run archive $A_t$,\\
    NSGA-II, MAP-Elites
};
\node[llmbox, right=1.2cm of controller] (llm) {
    \textbf{LLM mutation operator $q$}\\
    Hardware-specific prompt,\\
    Mutation / crossover
};
\node[evalbox, right=1.2cm of llm] (eval) {
    \textbf{Evaluate $f_i$}\\
    Full training,\\
    Train + energy,\\
    Robustness
};

\node[updatebox] (kb)
    at ($(llm.north)!0.5!(eval.north)+(0,1.2)$) {
    \textbf{Update}\\
    Knowledge Base $\mathcal{K}$
};
\node[updatebox] (pareto)
    at ($(controller.south)!0.5!(eval.south)+(0,-1.2)$) {
    \textbf{Update}\\
    Pareto front ${P}_i$
};

\draw[arrow] (inputs) -- (controller);
\draw[arrow] (controller) -- (llm);
\draw[arrow] (llm) -- (eval);

\draw[feedback] (eval.north) |- (kb.east);
\draw[feedback] (kb.west)    -| (llm.north);

\draw[feedback] (eval.south) |- (pareto.east);
\draw[feedback] (pareto.west) -| (controller.south);

\end{tikzpicture}%
}
\caption{
Overview of the UH-NAS framework. Each generation begins with hardware-specific inputs (sub-operation energy costs, seed architectures, and non-ideality parameters). A selection operator (NSGA-II with MAP-Elites diversity) chooses parents, which the LLM mutates or recombines via hardware-specific prompts. Candidates are trained and evaluated for accuracy, energy, and robustness under non-idealities. Resulting scores update the Pareto front and a knowledge base of design heuristics that guide subsequent LLM generations.
}
\label{fig:uhnas_schematic}
\end{figure*}

\begin{algorithm}[t]
\caption{UH-NAS: Hardware-Adaptive LLM-Guided NAS}
\label{alg:uhnas}
\small
\textbf{Input:} Hardware backend $\mathcal{B}_i$, 
dataset $\mathcal{D}$, generations $T$, candidates per gen $N$. \\
\textbf{Output:} \raggedright Pareto front $\mathcal{P}^*$ (accuracy $\uparrow$, energy $\downarrow$) 
\vspace{-0.36cm}
\begin{algorithmic}[1]
\STATE Initialize population $A_0$ from seed architectures.
\STATE $\mathcal{K} \gets \mathcal{B}_i.\textsc{DesignHeuristics}()$
  \hfill \textit{// knowledge base}
\FOR{$t = 0, 1, \dots, T$}
    \STATE $\mathcal{K} \gets \textsc{LLM-UpdateKB}(\mathcal{K},\;
      \textsc{ParetoFront}(A_t),\; \mathcal{B}_i)$
    \FOR{$k = 1, \dots, N$}
        \STATE $x' \gets \textsc{LLM-Generate}(
          \mathcal{K},\; \textsc{Select}(A_t),\; \mathcal{B}_i)$
        \STATE Score $x'$: $z \gets \textsc{LLM-Score}(x')$, \hfill \textit{// accuracy}
          $e \gets \mathcal{B}_i.\textsc{Energy}(x')$.
    \ENDFOR
    \STATE Keep top candidates by ($z \uparrow$, $e \downarrow$).
    \FOR{each surviving candidate $x$}
        \STATE $\mathrm{acc} \gets \textsc{Train}(x,\; \mathcal{B}_i,\; \mathcal{D})$
        \STATE $\mathrm{deg} \gets \mathrm{acc} - 
          \textsc{Eval}(\mathcal{B}_i.\textsc{InjectNoise}(x),\; \mathcal{D})$
        \STATE Insert $(x,\; \mathrm{acc},\; 
          \mathcal{B}_i.\textsc{Energy}(x),\; \mathrm{deg})$ 
          into $A_{t+1}$ via NSGA-II
    \ENDFOR
\ENDFOR
\STATE \textbf{return} $\mathcal{P}^* \gets \textsc{ParetoFront}(A_{t+1})$
\end{algorithmic}
\end{algorithm}

\subsubsection{Overall Framework}

UH-NAS serves as an iterative co-evolutionary framework that includes four stages per generation. Similar to prior NAS architectures \cite{ji2025rznas, zhu2025llmnas}, we integrate LLM-in-the-loop, utilizing it both as an evolutionary operator and a knowledge synthesizer that iteratively refines NAS design heuristics. Central to our architecture, UH-NAS is hardware-agnostic: all hardware-specific information (energy cost models, physical constraints, non-idealities, and quantization) are incorporated as replaceable hardware backend abstractions, allowing for easy comparisons between different hardware implementations without modifying the underlying NAS algorithm. These hardware considerations (Section \ref{sec:hardware_considerations}) allow for accurate estimation of model energy costs. We further integrate system prompts (imposing physical constraints found in certain hardware implementations), which we ablate in Table \ref{tab:nas_prompt_ablate}, along with following LLM-NAS's \cite{zhu2025llmnas} knowledge base (updatable architecture-specific rules updated every NAS generation that guide its search). 

The full pseudocode for UH-NAS is described in Algorithm~\ref{alg:uhnas}.


\subsubsection{Population Initialization}

The search space begins by instantiating a diverse set of hand-designed architectures $x_i^{(0)}$ that cover different parts of the search space (MLP, convolutional-linear, etc.). The different layers are described as typed blocks within an architectural vocabulary, similar to RZ-NAS \cite{ji2025rznas}, and are concatenated to form a block string that is validated by a forward pass. This vocabulary is updated depending on the hardware being tested (and this is passed in as part of the LLM generation prompt).

\subsubsection{Population Augmentation}

At each NAS generation, the LLM receives the global Pareto front, the recently evaluated candidates, and the accuracy degradation induced by hardware non-idealities. From these, the LLM is prompted to update the design heuristics as a structured list, allowing the LLM to discover robustness patterns not inherently visible from model performance. 

New candidate architectures are generated through mutation and crossover operations guided by this evolving knowledge base. Parents are selected from the current archive based on multi-objective fitness and diversity criteria, while the LLM proposes architectural modifications conditioned on prior search history. Formally, candidate generation follows
\begin{align}
x_i^{(t+1)} \sim q(\cdot \mid A_t, \mathcal{K}).
\end{align}
We denote by $q(\cdot \mid A_t,\mathcal{K})$ the LLM-induced proposal distribution over candidate architectures, conditioned on the current archive $A_t$ and the hardware-specific knowledge base $\mathcal{K}$. Then, for each candidate in a generation, a target region in a MAP-Elites diversity grid is sampled with a 70/30 exploit/explore ratio, similar to LLMatic \cite{nasir2024llmatic}. Parents are selected from the NSGA-II population via binary tournament \cite{Deb2002nsgaii}. This is then used to prompt the LLM, encouraging either crossover or mutation based on its priors to generate new candidates. More information on our multi-objective optimization is found in Section \ref{sec:multi_objective}.

For candidate architectures, we perform a full training loop consisting of 20-epoch training with Adam optimizer and cross-entropy loss, along with quantization-aware training (QAT) support for unconventional hardware with limited bit precision and non-ideality robustness calculations. The candidate is then inserted back into the NSGA-II population with information about its accuracy, energy, and robustness degradation.


\subsubsection{Multi-Objective Selection}\label{sec:multi_objective}


We leverage NSGA-II \cite{Deb2002nsgaii} for Pareto front generation, co-optimizing accuracy ($\uparrow$) and inference energy ($\downarrow$), although our framework supports arbitrary objective sets. Ranked Pareto fronts are generated, where rank-0 candidates are not dominated by any other individual, rank-1 candidates are dominated solely by rank-0 members, etc. Crowding distances are used to quantify the isolation of each candidate, encouraging spread across the front.

For parent selection, a binary tournament is used where two candidates are drawn at random, with the lower Pareto rank chosen as a parent. For crossover, two tournaments produce two parents while for mutation, a single tournament suffices. After candidates are evaluated and inserted, the population is truncated, yielding the output: a final rank-0 Pareto front representing the set of non-dominated accuracy/energy tradeoffs.

Given the poor correlation between ZC methods and non-ideality accuracy observed in Figure~\ref{fig:nb201_zc_spearman_worst_case}, we decide to train all architectures from scratch (as in LLMatic \cite{nasir2024llmatic}). We posit this slight negative correlation between non-ideal accuracy and ZC proxies can lead to destructive performance in NAS for MZI-based optical neural networks (ONNs). 

\subsection{Hardware Considerations}\label{sec:hardware_considerations}

We model each hardware backend through an energy cost function and, for the optical regime, an additional structured non-ideality model. Table \ref{tab:energy_estimates} summarizes per-operation energy estimates across CPU (Xeon 8380, FP32), GPU (B200, FP16), and the MZI-based ONN (8-bit precision). Linear MAC costs span roughly four orders of magnitude and are significantly cheaper on optical hardware, with the tradeoff being additional DAC/ADC conversion costs at network inputs and outputs that partially offset per-MAC advantage. Exact derivations for each are included in Appendix \ref{app:energy_efficiency}.

\begin{table}[h]
\centering
\begin{tabular}{lccc}
\toprule
 & \textbf{CPU} & \textbf{GPU} & \textbf{Optical} \\
 & (Xeon 8380) & (B200) & (MZI) \\
\midrule
Linear MAC          & 91.7 pJ      & 0.89 pJ      & 0.02 pJ \\
Nonlinear (ReLU)    & 3 pJ        & 3 pJ         & 10 pJ \\
DAC (input)         & \diagbox{}{} & \diagbox{}{} & 2 pJ \\
ADC (output)        & \diagbox{}{} & \diagbox{}{} & 4 pJ \\
\midrule
Precision (bits)    & 32 (FP32)    & 16 (FP16)    & 8 \\
\bottomrule
\end{tabular}
\caption{Per-operation analytical energy proxy coefficients for CPU, GPU, and optical (MZI ONN) inference. Linear MAC is per multiply-accumulate; nonlinear, DAC, and ADC are per element. DAC/ADC are not applicable to digital CPU/GPU inference.}
\label{tab:energy_estimates}
\end{table}

For the optical backend, we further capture three physical noise sources: phase error ($\sigma_\phi$), thermal crosstalk ($\varepsilon$), and gamma noise ($\sigma_\gamma$) in Table \ref{tab:nonidealities}. Realistic and worst-case noise values correspond to approximate lower and upper bounds reported in silicon-photonic literature, while quantization noise is handled separately via QAT for 8-bit representations that are commonplace in ONN literature. Full derivations of the energy model, the architectural constraints imposed on MZI meshes (block-diagonal decomposition, width caps, residual-connection penalties), and detailed descriptions of each noise source are provided in Appendix \ref{app:non_idealities}.

\begin{table}[h]
\centering
\begin{tabular}{lccc}
\toprule
 & \textbf{Phase error} & \textbf{Crosstalk} & \textbf{Gamma noise} \\
 & ($\sigma_\phi$) & ($\varepsilon$) & ($\sigma_\gamma$) \\
\midrule
Realistic    & 0.01  & 0.01  & 0.001 \\
Medium  & 0.03  & 0.08  & 0.003 \\
Worst   & 0.05  & 0.15  & 0.005 \\
\bottomrule
\end{tabular}
\caption{Hardware non-ideality levels evaluated for optical regime. Realistic-case and worst-case values correspond to the lower and upper bounds reported in literature; medium-case is the midpoint.}
\label{tab:nonidealities}
\end{table}


\section{Experiments}

\subsection{Experimental Setup}

We evaluate UH-NAS on three hardware backends: a silicon-photonics MZI-mesh backend (simulated via \texttt{pytorch-onn}), a GPU reference backend (Blackwell on FP16), and a CPU reference backend (Xeon on FP32) tested on MNIST. The search procedure runs for 30 generations with 8 candidate architectures generated per generation. The NSGA-II population size is set to 20. Each candidate is trained for 20 epochs using Adam (lr=$10^{-3}$) with cross-entropy loss. Seed populations consist of 10 hand-designed architectures spanning pure MLP, convolutional-linear hybrid, and varying-depth configurations.

For competing architectures (LLMatic, RZ-NAS), we leverage the same seed architectures and optimize for non-ideality accuracy to encourage fairer comparisons, along with updating the search space to utilize the same MZI-typed vocabulary. For LLMatic, we utilize the same training procedure (30 generations, 8 candidate architectures per generation trained under the same conditions), maintaining the MAP-Elites archive and algorithm (same mutation and crossover prompts, archive sizes, curiosity/temperature update rules, etc.), switching the fitness signal to non-ideality accuracy for non-ideal LLMatic. For RZ-NAS we evaluate 1500 generations with a population size of 100 and utilize SynFlow as the zero-cost fitness function, also reporting non-ideality accuracy on the final-trained winner. We apply no additional constraints on parameter count/model size, but still utilize QAT for training and implement the same MZI ONN architectural limitations.

\begin{table*}[!htbp]
\caption{Top-1 MNIST accuracy under worst-case optical non-idealities for UH-NAS and baselines (RZ-NAS, LLMatic), each using three LLM backbones. All methods share the same seed architectures and optimize for non-ideality accuracy. UH-NAS with GPT-4.1 achieves the highest mean accuracy (97.00\%), outperforming both baselines and the ablation without LLM-guided generation.
}
\label{tab:results-nas}
\begin{center}
\vskip -0.06in
\begin{small}
\begin{tabular}{ lccc } 
 \toprule
 \textbf{Method} & \textbf{GPT-4.1 Nano} & \textbf{GPT-4.1} & \textbf{Gemini 3.1 Flash-Lite}\\ 
 \midrule
 UH-NAS & 91.96\% $\pm$ 0.43\% & \textbf{97.00\% $\pm$ 0.30\%} & \textbf{96.95\% $\pm$ 0.03\%} \\ 
 UH-NAS (w/o LLM) & 92.74\% $\pm$ 0.18\% & 92.46\% $\pm$ 0.35\% & 92.61\% $\pm$ 0.31\% \\  
 \midrule
 RZ-NAS \cite{ji2025rznas} &  46.42\% $\pm$ 20.71\% & 88.69\% $\pm$ 6.85\% & 19.16\% $\pm$ 8.11\% \\ 
 LLMatic \cite{nasir2024llmatic} & 47.07\% $\pm$ 38.00\% & 66.03\% $\pm$ 14.51\% & 51.93\% $\pm$ 23.70\% \\ 
 LLMatic (non-ideal)  & \textbf{93.40\% $\pm$ 1.45} \% & 91.81\% $\pm$ 1.83\% & 92.92\% $\pm$ 2.09\% \\ 
 \bottomrule
\end{tabular}
\end{small}
\end{center}
\vskip -0.15in
\end{table*}

\subsection{Cross-Platform Pareto Analysis: Summary}

Figure~\ref{fig:architecture} summarizes UH-NAS search outcomes across GPU, CPU, and MZI-based optical backends through their respective Pareto-optimal accuracy–energy trade-offs. A key advantage of the framework is enabling system-level, fair comparison across hardware platforms, as architectures are optimized independently for each hardware under its specific constraints and non-idealities. GPU architectures achieve significantly lower energy per inference than CPU architectures at comparable validation error. In contrast, optical hardware operates in a substantially lower energy regime, with a moderate accuracy degradation under non-idealities. UH-NAS reduces this gap by identifying architectures that improve robustness, indicating that optical platforms can approach competitive accuracy while retaining strong energy-efficiency advantages.

\subsection{Benefits of LLM-Guided Search for Hardware-Aware NAS}

\begin{figure}[t]
  \centering
\includegraphics[width=\columnwidth]{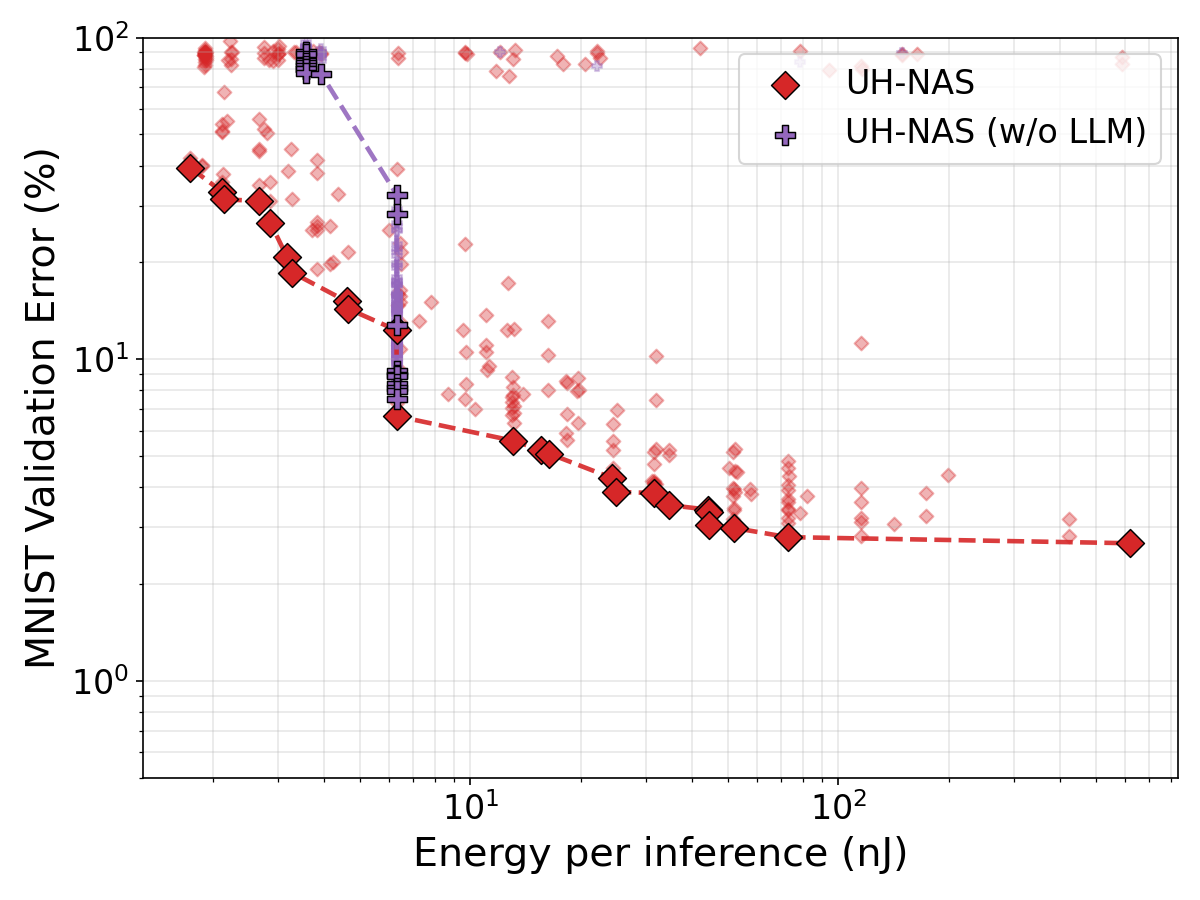}
    \caption{Ablation comparing UH-NAS (with LLM) and UH-NAS without LLM on the MZI optical backend under worst-case non-idealities. Without LLM-guided generation, the search degenerates to minor seed architecture modifications, producing only 26 unique designs out of 250 evaluations and a restricted Pareto front. With the LLM, UH-NAS explores 203 unique architectures, yielding broader coverage and lower validation error at comparable energy budgets.
    }
  \label{fig:search-diversity}
\end{figure}

In Table~\ref{tab:results-nas}, compared to RZ-NAS and LLMatic, we note that UH-NAS tends to obtain higher top accuracy on MNIST (Table~\ref{tab:results-nas}), along with being capable of co-optimizing both accuracy and efficiency (while other algorithms focus solely on performance thus leading to very deep searched architectures). We posit this advantage stems from UH-NAS's hardware awareness, driven in part by the LLM prompt guiding the algorithm. We observe that UH-NAS with a strong LLM backbone noticeably outperforms existing baselines, even when these baselines are explicitly trained/optimized for non-ideal accuracy (e.g. LLMatic with non-ideality accuracy as a criterion). This LLM backbone plays a key role in the system prompt/knowledge base generation, as weaker models such as GPT-4.1-nano tend to introduce weak or incorrect prompts that degrade search performance, while stronger models leverage the structured context to achieve large gains (Appendix \ref{app:system_prompt}). We note that the poor performance of Gemini 3.1 Flash-Lite without system prompts is induced by the fact that it tends to bias towards convolutional layers, which exhibit catastrophic performance degradation under non-ideality conditions. 

Figure~\ref{fig:search-diversity} shows that UH-NAS without the guiding LLM rapidly collapses to a narrow region of the search space, yielding limited architectural diversity (26 valid unique architectures out of 250 evaluations) and a restricted Pareto front. In contrast, UH-NAS explores a substantially broader design space (203 valid unique architectures), resulting in improved coverage and lower validation error at comparable energy. This indicates that LLM-guided generation enables non-local exploration and mitigates search degeneracy induced by hardware constraints.


\subsection{Robust Architecture Discovery Under Non-Idealities}

\begin{figure*}[t]
  \centering
\includegraphics[width=0.9\linewidth]{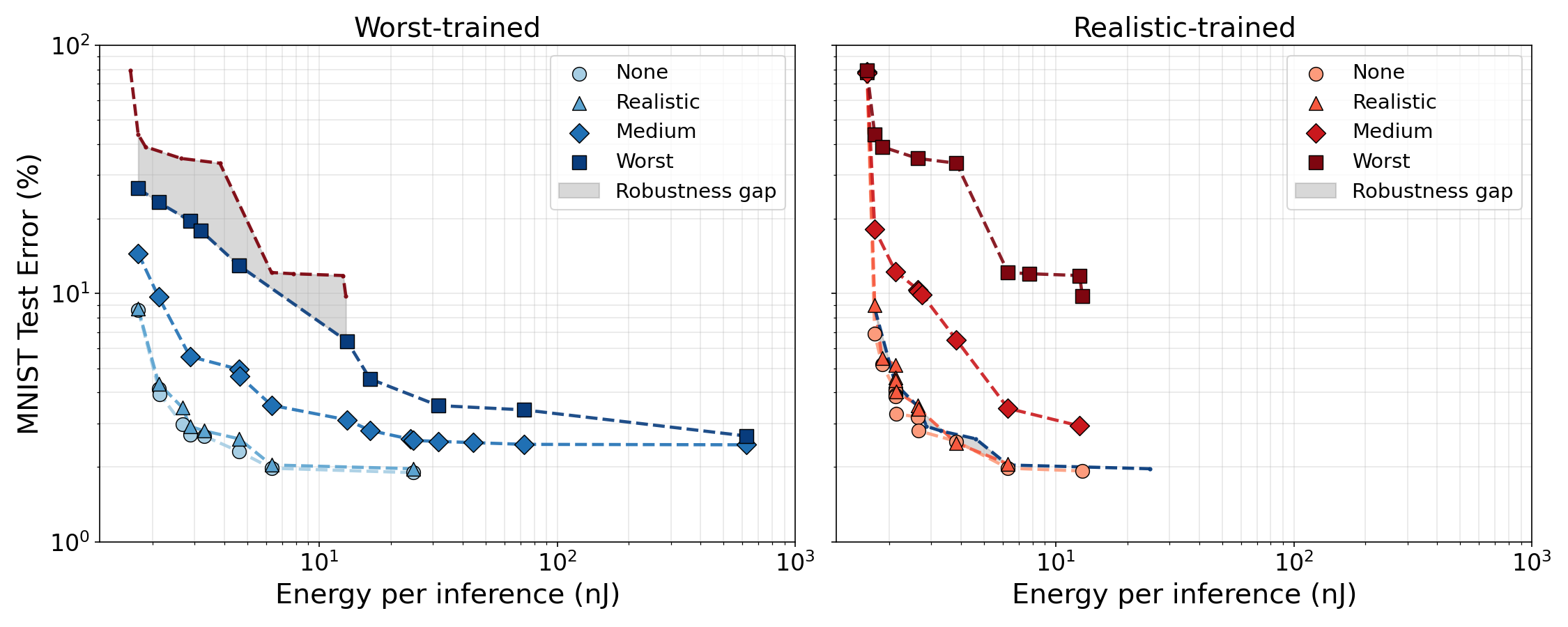}
    \caption{Robustness of UH-NAS–discovered architectures across non-ideality regimes on the MZI optical backend. The Pareto front optimized under worst-case noise (left) and realistic noise (right) is re-evaluated at four non-ideality levels: none, realistic, medium, and worst-case, with the shaded gray region highlighting the robustness gap of the other training environment (worst-case comparison on the left, realistic on the right). In both panels, the Pareto front structure is preserved across regimes, with a consistent upward shift in error as noise increases. Worst-case–trained architectures remain competitive even under moderate and realistic conditions, suggesting that robustness generalizes beyond the specific noise level used during search.
    }
  \label{fig:TBA2}
\end{figure*}
\begin{figure*}[t]
    \centering
    \includegraphics[width=0.9\textwidth]{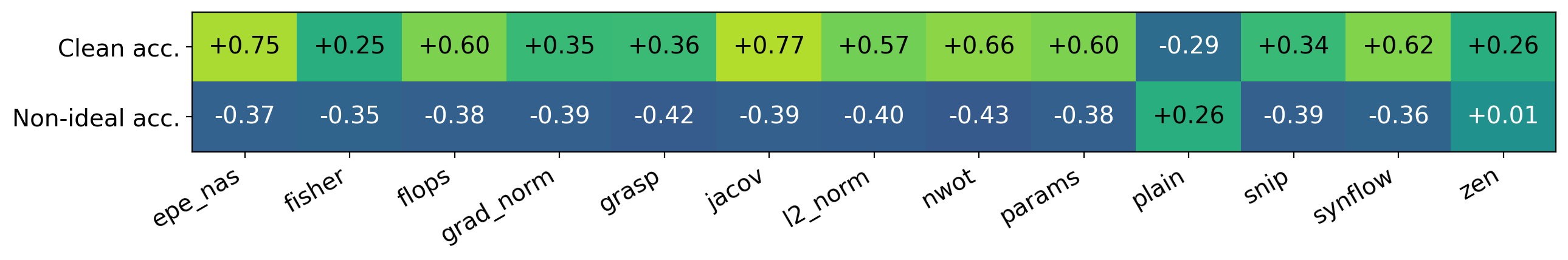}
    \caption{Spearman rank correlation between 13 zero-cost (ZC) proxies and validation accuracy on 200 randomly drawn NAS-Bench-201/CIFAR-10 architectures, evaluated under clean and worst-case MZI ONN non-idealities ($\sigma_\phi=0.05$, $\sigma_\gamma=0.005$, crosstalk$=0.15$). Under clean conditions, most proxies show positive correlations consistent with prior literature. Under worst-case non-idealities, nearly all proxies exhibit negative correlations, indicating that ZC proxies are unreliable for NAS on non-ideal optical hardware.}
    \label{fig:nb201_zc_spearman_worst_case}
\end{figure*}

Figure~\ref{fig:TBA2} evaluates the robustness of architectures discovered under worst-case non-idealities by re-evaluating the same Pareto front across varying noise regimes. Performance degrades with increasing non-idealities, but the Pareto front remains well-structured, indicating that architectures discovered under worst-case conditions retain strong performance across all regimes (while the inverse is not necessarily true). Notably, the worst-case-trained architectures remain competitive even under moderate and realistic non-idealities, suggesting that robustness learned during search generalizes beyond the specific noise level used for optimization.

Figure~\ref{fig:nb201_zc_spearman_worst_case} reports Spearman correlations on 200 NAS-Bench-201 architectures under both clean and worst-case MZI ONN conditions. Under clean conditions, we observe correlations in line with prior ZC NAS literature \cite{ji2025rznas, krishnakumar2022nasbenchsuitezero, zhu2025llmnas}, but under non-ideality conditions nearly every proxy exhibits negative correlations, mostly induced by the destructive performance of convolutional layers under worst-case non-idealities (which typically exhibit strong positive correlations with ZC methods). These findings justify UH-NAS's design decision to score every candidate via full training, while explaining the poor performance of ZC proxy NAS methods observed in Table~\ref{tab:results-nas}.

\subsection{Ablation Studies}

\begin{table*}[!htbp]
\caption{Ablation on the role of the hardware-informed system prompt ($\mathcal{K}$) in UH-NAS under worst-case optical non-idealities. Removing the system prompt reduces accuracy across strong LLM backbones (GPT-4.1, Gemini 3.1 Flash-Lite), indicating that structured hardware context is essential for effective LLM-guided search given accurate construction of the prompt.}
\label{tab:nas_prompt_ablate}
\begin{center}
\vskip -0.06in
\begin{small}
\begin{tabular}{ lccc } 
 \toprule
 \textbf{Method} & \textbf{GPT-4.1 Nano} & \textbf{GPT-4.1} & \textbf{Gemini 3.1 Flash-Lite}\\ 
 \midrule
 UH-NAS & 91.96\% $\pm$ 0.43\% & \textbf{97.00\% $\pm$ 0.30\%} & \textbf{96.95\% $\pm$ 0.03\%} \\ 
 UH-NAS (w/o $\mathcal{K}$) & \textbf{93.21\% $\pm$ 2.34\%} & 96.02\% $\pm$ 1.59\% & 96.55\% $\pm$ 0.42\% \\ 
 \bottomrule
\end{tabular}
\end{small}
\end{center}
\vskip -0.15in
\end{table*}


Table~\ref{tab:nas_prompt_ablate} compares UH-NAS with and without the hardware-informed knowledge base ($\mathcal{K}$) under worst-case non-idealities. The results show that incorporating structured context about the hardware improves performance for stronger LLMs (e.g., GPT-4.1 and Gemini 3.1 Flash-Lite), highlighting the importance of hardware-aware prompting and contextual guidance during search.

\begin{figure}[t]
  \centering
\includegraphics[width=\columnwidth]{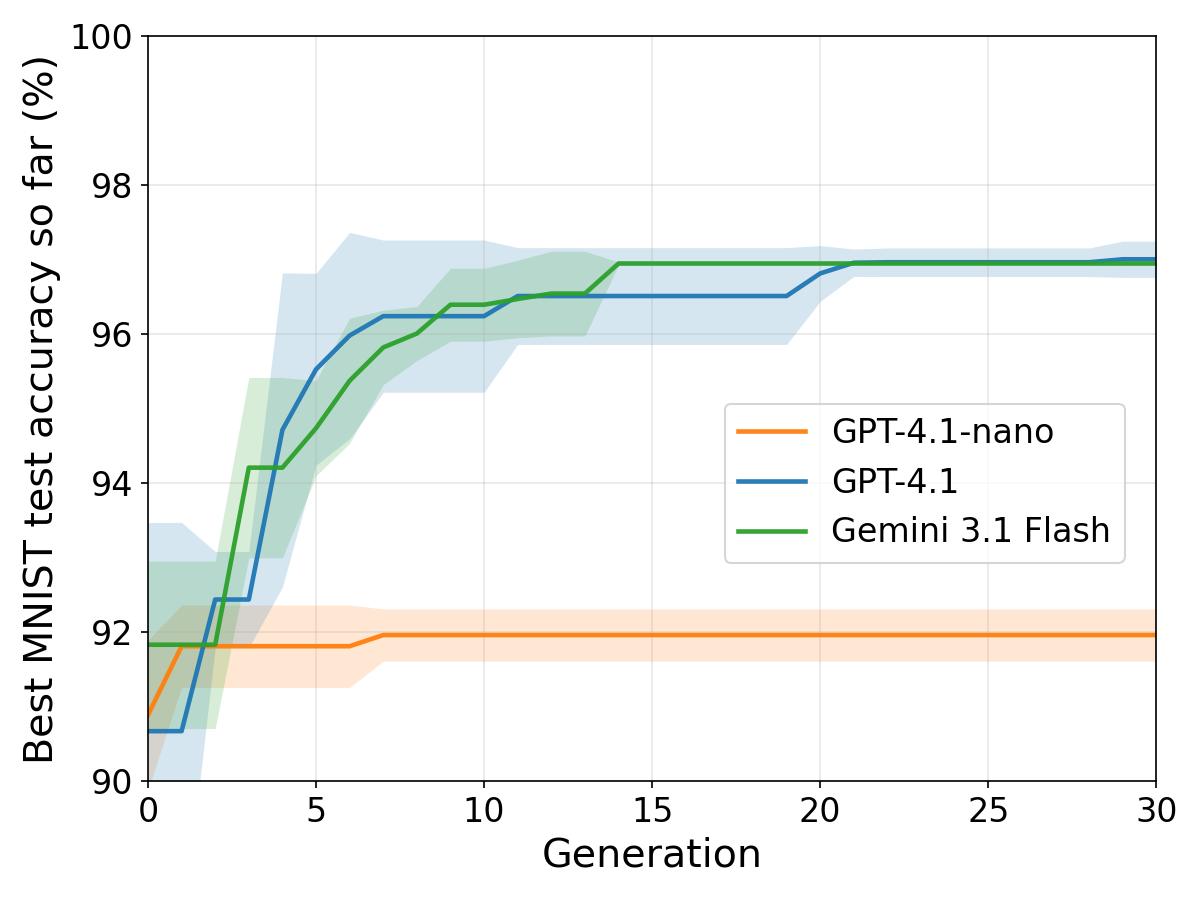}
    \caption{Best observed MNIST accuracy (worst-case non-idealities) over 30 NAS generations, averaged over 3 runs, for three LLM backbones. GPT-4.1 and Gemini 3.1 Flash-Lite converge to approximately 97\% accuracy within 10–15 generations, while GPT-4.1-nano shows limited improvement due to its reduced capacity for generating effective system prompts and diverse architectures. }
  \label{fig:accuracy_generation}
\end{figure}

Figure~\ref{fig:accuracy_generation} compares the evolution of best accuracy across generations for different LLMs. Stronger models (GPT-4.1 and Gemini 3.1 Flash-Lite) achieve both faster convergence and higher final accuracy compared to smaller models (GPT-4.1 nano), which shows little to no improvement over iterations. This indicates that LLM capability directly impacts the effectiveness of the search process, with more capable models enabling non-trivial exploration and more efficient identification of high-performing architectures.

\subsection{Discovered Architectures}

On both CPU and GPU backends, we observe that high-accuracy Pareto-optimal architectures tend to follow traditional machine learning design principles, operating as convolutional-linear hybrid patterns with 4-6 layers (akin to LeNet-5 \cite{Lecun1998lenet}) that exploit depth to maximize representational capacity. However, on the optical backend we observe a different trend: shallow, wide MLPs tend to dominate, with convolutional layers largely absent outside of the energy-efficient end of the Pareto front. In all settings, normalization frequently appears at high-energy regimes, and is particularly helpful for settings with worst-case non-idealities. We observe no skip connections in the optical Pareto front, likely due to the large overhead costs associated with signal re-routing and additional ADC/DAC conversions.

Such hardware differences are physically grounded. Depth sensitivity arises due to phase error ($\sigma_\phi$) compounding through sequential MZI mesh layers. Convolutional layers fail catastrophically due to crosstalk, which propagates weight-shifted convolutional kernel errors across the output feature map \cite{Su2024reliability}. Batch normalization, by comparison, operates as a noise recalibration mechanism, re-centering activations after noisy operations introduced by phase errors and crosstalk. Finally, the lack of skip connections and nonlinearities occur due to the extreme overhead from ADC/DAC costs, which are $\geq2$ orders of magnitude more expensive than optical MACs.

These findings provide a set of actionable design principles for MZI-based ONN that can help guide future architectures: namely that shallow, wide networks are preferred, utilizing solely linear layers under high crosstalk and integrating batch normalization to mitigate non-ideality model drift. These principles highlight a key feature of UH-NAS: the discovery of novel, robust architectures across heterogeneous hardware platforms. Compared to traditional NAS methods, our approach discovers a multitude of optimal architectures at varying energy budgets, avoiding common pitfalls such as overweighting deep architectures that may not be as resistant to hardware non-idealities or the utilization of ZC proxies that incorrectly score highest performers.


\section{Conclusion}

We presented UH-NAS, an LLM-guided NAS framework for unconventional hardware that co-optimizes accuracy and inference energy across heterogeneous platforms via swappable backend abstractions, enabling fair system-level comparisons. We integrate LLM feedback to guide the NAS, constraining the search space and iteratively delivering hardware-aware feedback. On MZI-based optical hardware under worst-case non-idealities, UH-NAS achieves 97\% MNIST accuracy while discovering more diverse architectures compared to conventional baselines, with Pareto fronts that remain well-structured across noise regimes.

Beyond framework performance, UH-NAS uncovers physically interpretable design principles for ONNs: shallow, wide MLPs outperform convolutional-linear hybrids due to multiplicative phase noise compounding through depth and correlated crosstalk inducing errors that occur in convolutional weight reuse. Furthermore, batch normalization acts as a noise recalibration mechanism and skip connections are absent due to prohibitive ADC/DAC conversion costs. We hope these principles guide future neural networks deployed on photonic accelerators, bridging the gap between automated search and physical hardware. 

Several limitations remain, however. UH-NAS assumes feedforward architectures composed of standard layers, which may not extend to fundamentally different hardware such as spiking neuromorphic processors or quantum computing. Additionally, while the framework is algorithmically hardware-agnostic, it still requires hardware-specific inputs, including seed architectures, energy models, and non-ideality simulators. Natural extensions include scaling to harder datasets, expanding to additional unconventional computing paradigms (e.g. memristors), and reducing computational costs through early-stopping strategies such as Hyperband-style successive halving \cite{li2018hyperband}.

\section*{Impact Statement}

This paper presents work whose goal is to advance the field of Machine
Learning. There are many potential societal consequences of our work, none
which we feel must be specifically highlighted here.

\bibliographystyle{unsrt}
\bibliography{lib}

@article{ye2024reevo,
  title={Reevo: Large language models as hyper-heuristics with reflective evolution},
  author={Ye, Haoran and Wang, Jiarui and Cao, Zhiguang and Berto, Federico and Hua, Chuanbo and Kim, Haeyeon and Park, Jinkyoo and Song, Guojie},
  journal={Advances in neural information processing systems},
  volume={37},
  pages={43571--43608},
  year={2024}
}

@article{liu2024evolution,
  title={Evolution of heuristics: Towards efficient automatic algorithm design using large language model},
  author={Liu, Fei and Tong, Xialiang and Yuan, Mingxuan and Lin, Xi and Luo, Fu and Wang, Zhenkun and Lu, Zhichao and Zhang, Qingfu},
  journal={arXiv preprint arXiv:2401.02051},
  year={2024}
}

@article{novikov2025alphaevolve,
  title={Alphaevolve: A coding agent for scientific and algorithmic discovery},
  author={Novikov, Alexander and V{\~u}, Ng{\^a}n and Eisenberger, Marvin and Dupont, Emilien and Huang, Po-Sen and Wagner, Adam Zsolt and Shirobokov, Sergey and Kozlovskii, Borislav and Ruiz, Francisco JR and Mehrabian, Abbas and others},
  journal={arXiv preprint arXiv:2506.13131},
  year={2025}
}

@article{romera2024mathematical,
  title={Mathematical discoveries from program search with large language models},
  author={Romera-Paredes, Bernardino and Barekatain, Mohammadamin and Novikov, Alexander and Balog, Matej and Kumar, M Pawan and Dupont, Emilien and Ruiz, Francisco JR and Ellenberg, Jordan S and Wang, Pengming and Fawzi, Omar and others},
  journal={Nature},
  volume={625},
  number={7995},
  pages={468--475},
  year={2024},
  publisher={Nature Publishing Group UK London}
}

@article{fawzi2022discovering,
  title={Discovering faster matrix multiplication algorithms with reinforcement learning},
  author={Fawzi, Alhussein and Balog, Matej and Huang, Aja and Hubert, Thomas and Romera-Paredes, Bernardino and Barekatain, Mohammadamin and Novikov, Alexander and R. Ruiz, Francisco J and Schrittwieser, Julian and Swirszcz, Grzegorz and others},
  journal={Nature},
  volume={610},
  number={7930},
  pages={47--53},
  year={2022},
  publisher={Nature Publishing Group UK London}
}

@article{bessalah2025analognas,
  title={AnalogNAS-Bench: A NAS Benchmark for Analog In-Memory Computing},
  author={Bessalah, Aniss and Abdelmoumen, Hatem Mohamed and Benatchba, Karima and Benmeziane, Hadjer},
  journal={arXiv preprint arXiv:2506.18495},
  year={2025}
}

@inproceedings{negi2022nax,
  title={NAX: neural architecture and memristive xbar based accelerator co-design},
  author={Negi, Shubham and Chakraborty, Indranil and Ankit, Aayush and Roy, Kaushik},
  booktitle={Proceedings of the 59th ACM/IEEE Design Automation Conference},
  pages={451--456},
  year={2022}
}

@article{zhou2021analognets,
  title={AnalogNets: ML-HW co-design of noise-robust TinyML models and always-on analog compute-in-memory accelerator},
  author={Zhou, Chuteng and Redondo, Fernando Garcia and B{\"u}chel, Julian and Boybat, Irem and Comas, Xavier Timoneda and Nandakumar, SR and Das, Shidhartha and Sebastian, Abu and Gallo, Manuel Le and Whatmough, Paul N},
  journal={arXiv preprint arXiv:2111.06503},
  year={2021}
}

@article{yuan2021nas4rram,
  title={NAS4RRAM: neural network architecture search for inference on RRAM-based accelerators},
  author={Yuan, Zhihang and Liu, Jingze and Li, Xingchen and Yan, Longhao and Chen, Haoxiang and Wu, Bingzhe and Yang, Yuchao and Sun, Guangyu},
  journal={Science China Information Sciences},
  volume={64},
  number={6},
  pages={160407},
  year={2021},
  publisher={Springer}
}

@article{jiang2020device,
  title={Device-circuit-architecture co-exploration for computing-in-memory neural accelerators},
  author={Jiang, Weiwen and Lou, Qiuwen and Yan, Zheyu and Yang, Lei and Hu, Jingtong and Hu, Xiaobo Sharon and Shi, Yiyu},
  journal={IEEE Transactions on Computers},
  volume={70},
  number={4},
  pages={595--605},
  year={2020},
  publisher={IEEE}
}

@article{li2021flash,
  title={FLASH: F ast Neura l A rchitecture S earch with H ardware Optimization},
  author={Li, Guihong and Mandal, Sumit K and Ogras, Umit Y and Marculescu, Radu},
  journal={ACM Transactions on Embedded Computing Systems (TECS)},
  volume={20},
  number={5s},
  pages={1--26},
  year={2021},
  publisher={ACM New York, NY}
}

@article{zheng2023can,
  title={Can gpt-4 perform neural architecture search?},
  author={Zheng, Mingkai and Su, Xiu and You, Shan and Wang, Fei and Qian, Chen and Xu, Chang and Albanie, Samuel},
  journal={arXiv preprint arXiv:2304.10970},
  year={2023}
}

@article{chen2023evoprompting,
  title={Evoprompting: Language models for code-level neural architecture search},
  author={Chen, Angelica and Dohan, David and So, David},
  journal={Advances in neural information processing systems},
  volume={36},
  pages={7787--7817},
  year={2023}
}

@article{shastri2021photonics,
  title={Photonics for artificial intelligence and neuromorphic computing},
  author={Shastri, Bhavin J and Tait, Alexander N and Ferreira de Lima, Thomas and Pernice, Wolfram HP and Bhaskaran, Harish and Wright, C David and Prucnal, Paul R},
  journal={Nature Photonics},
  volume={15},
  number={2},
  pages={102--114},
  year={2021},
  publisher={Nature Publishing Group UK London}
}

@inproceedings{gu2022adept,
  title={Adept: Automatic differentiable design of photonic tensor cores},
  author={Gu, Jiaqi and Zhu, Hanqing and Feng, Chenghao and Jiang, Zixuan and Liu, Mingjie and Zhang, Shuhan and Chen, Ray T and Pan, David Z},
  booktitle={Proceedings of the 59th ACM/IEEE Design Automation Conference},
  pages={937--942},
  year={2022}
}

@inproceedings{wang2022quantumnas,
  title={Quantumnas: Noise-adaptive search for robust quantum circuits},
  author={Wang, Hanrui and Ding, Yongshan and Gu, Jiaqi and Lin, Yujun and Pan, David Z and Chong, Frederic T and Han, Song},
  booktitle={2022 IEEE International Symposium on High-Performance Computer Architecture (HPCA)},
  pages={692--708},
  year={2022},
  organization={IEEE}
}

@inproceedings{yan2021uncertainty,
  title={Uncertainty modeling of emerging device based computing-in-memory neural accelerators with application to neural architecture search},
  author={Yan, Zheyu and Juan, Da-Cheng and Hu, Xiaobo Sharon and Shi, Yiyu},
  booktitle={Proceedings of the 26th Asia and South Pacific Design Automation Conference},
  pages={859--864},
  year={2021}
}

@article{bhattacharjee2023xplorenas,
  title={XploreNAS: Explore adversarially robust and hardware-efficient neural architectures for non-ideal xbars},
  author={Bhattacharjee, Abhiroop and Moitra, Abhishek and Panda, Priyadarshini},
  journal={ACM Transactions on Embedded Computing Systems},
  volume={22},
  number={4},
  pages={1--17},
  year={2023},
  publisher={ACM New York, NY}
}

@inproceedings{benmeziane2023analognas,
  title={Analognas: A neural network design framework for accurate inference with analog in-memory computing},
  author={Benmeziane, Hadjer and Lammie, Corey and Boybat, Irem and Rasch, Malte and Le Gallo, Manuel and Tsai, Hsinyu and Muralidhar, Ramachandran and Niar, Smail and Hamza, Ouarnoughi and Narayanan, Vijay and others},
  booktitle={2023 IEEE International Conference on Edge Computing and Communications (EDGE)},
  pages={233--244},
  year={2023},
  organization={IEEE}
}

@article{li2021hw,
  title={Hw-nas-bench: Hardware-aware neural architecture search benchmark},
  author={Li, Chaojian and Yu, Zhongzhi and Fu, Yonggan and Zhang, Yongan and Zhao, Yang and You, Haoran and Yu, Qixuan and Wang, Yue and Lin, Yingyan Celine},
  journal={arXiv preprint arXiv:2103.10584},
  year={2021}
}

@inproceedings{tan2019mnasnet,
  title={Mnasnet: Platform-aware neural architecture search for mobile},
  author={Tan, Mingxing and Chen, Bo and Pang, Ruoming and Vasudevan, Vijay and Sandler, Mark and Howard, Andrew and Le, Quoc V},
  booktitle={Proceedings of the IEEE/CVF conference on computer vision and pattern recognition},
  pages={2820--2828},
  year={2019}
}

@article{elsken2019neural,
  title={Neural architecture search: A survey},
  author={Elsken, Thomas and Metzen, Jan Hendrik and Hutter, Frank},
  journal={Journal of Machine Learning Research},
  volume={20},
  number={55},
  pages={1--21},
  year={2019}
}

@misc{zhu2025llmnas,
      title={LLM-NAS: LLM-driven Hardware-Aware Neural Architecture Search}, 
      author={Hengyi Zhu and Grace Li Zhang and Shaoyi Huang},
      year={2025},
      eprint={2510.01472},
      archivePrefix={arXiv},
      primaryClass={cs.LG},
      url={https://arxiv.org/abs/2510.01472}, 
}

@article{Shen2017nanophotonics,
  title = {Deep learning with coherent nanophotonic circuits},
  volume = {11},
  ISSN = {1749-4893},
  url = {http://dx.doi.org/10.1038/nphoton.2017.93},
  DOI = {10.1038/nphoton.2017.93},
  number = {7},
  journal = {Nature Photonics},
  publisher = {Springer Science and Business Media LLC},
  author = {Shen,  Yichen and Harris,  Nicholas C. and Skirlo,  Scott and Prabhu,  Mihika and Baehr-Jones,  Tom and Hochberg,  Michael and Sun,  Xin and Zhao,  Shijie and Larochelle,  Hugo and Englund,  Dirk and Soljačić,  Marin},
  year = {2017},
  month = june,
  pages = {441–446}
}

@inproceedings{jiaqigu2021L2ight,
    title     = {L2ight: Enabling On-Chip Learning for Optical Neural Networks via Efficient in-situ Subspace Optimization},
    author    = {Jiaqi Gu and Hanqing Zhu and Chenghao Feng and Zixuan Jiang and Ray T. Chen and David Z. Pan},
    booktitle = {Conference on Neural Information Processing Systems (NeurIPS)},
    year      = {2021}
}

@article{Demirkiran2023electrophotonic,
  title = {An Electro-Photonic System for Accelerating Deep Neural Networks},
  volume = {19},
  ISSN = {1550-4840},
  url = {http://dx.doi.org/10.1145/3606949},
  DOI = {10.1145/3606949},
  number = {4},
  journal = {ACM Journal on Emerging Technologies in Computing Systems},
  publisher = {Association for Computing Machinery (ACM)},
  author = {Demirkiran,  Cansu and Eris,  Furkan and Wang,  Gongyu and Elmhurst,  Jonathan and Moore,  Nick and Harris,  Nicholas C. and Basumallik,  Ayon and Reddi,  Vijay Janapa and Joshi,  Ajay and Bunandar,  Darius},
  year = {2023},
  month = sept,
  pages = {1–31}
}

@article{Vadlamani2023transferable,
  title = {Transferable learning on analog hardware},
  volume = {9},
  ISSN = {2375-2548},
  url = {http://dx.doi.org/10.1126/sciadv.adh3436},
  DOI = {10.1126/sciadv.adh3436},
  number = {28},
  journal = {Science Advances},
  publisher = {American Association for the Advancement of Science (AAAS)},
  author = {Vadlamani,  Sri Krishna and Englund,  Dirk and Hamerly,  Ryan},
  year = {2023},
  month = july 
}

@article{Hamerly2022scalable,
   title={Asymptotically fault-tolerant programmable photonics},
   volume={13},
   ISSN={2041-1723},
   url={http://dx.doi.org/10.1038/s41467-022-34308-3},
   DOI={10.1038/s41467-022-34308-3},
   number={1},
   journal={Nature Communications},
   publisher={Springer Science and Business Media LLC},
   author={Hamerly, Ryan and Bandyopadhyay, Saumil and Englund, Dirk},
   year={2022},
   month=Nov 
}

@article{Pai2023backprop,
  title = {Experimentally realized in situ backpropagation for deep learning in photonic neural networks},
  volume = {380},
  ISSN = {1095-9203},
  url = {http://dx.doi.org/10.1126/science.ade8450},
  DOI = {10.1126/science.ade8450},
  number = {6643},
  journal = {Science},
  publisher = {American Association for the Advancement of Science (AAAS)},
  author = {Pai,  Sunil and Sun,  Zhanghao and Hughes,  Tyler W. and Park,  Taewon and Bartlett,  Ben and Williamson,  Ian A. D. and Minkov,  Momchil and Milanizadeh,  Maziyar and Abebe,  Nathnael and Morichetti,  Francesco and Melloni,  Andrea and Fan,  Shanhui and Solgaard,  Olav and Miller,  David A. B.},
  year = {2023},
  month = apr,
  pages = {398–404}
}

@article{Milanizadeh2019canceling,
  title = {Canceling Thermal Cross-Talk Effects in Photonic Integrated Circuits},
  volume = {37},
  ISSN = {1558-2213},
  url = {http://dx.doi.org/10.1109/JLT.2019.2892512},
  DOI = {10.1109/jlt.2019.2892512},
  number = {4},
  journal = {Journal of Lightwave Technology},
  publisher = {Institute of Electrical and Electronics Engineers (IEEE)},
  author = {Milanizadeh,  Maziyar and Aguiar,  Douglas and Melloni,  Andrea and Morichetti,  Francesco},
  year = {2019},
  month = feb,
  pages = {1325–1332}
}

@inproceedings{Gu2020roq,
  title = {ROQ: A Noise-Aware Quantization Scheme Towards Robust Optical Neural Networks with Low-bit Controls},
  url = {http://dx.doi.org/10.23919/DATE48585.2020.9116521},
  DOI = {10.23919/date48585.2020.9116521},
  booktitle = {2020 Design,  Automation; Test in Europe Conference; Exhibition (DATE)},
  publisher = {IEEE},
  author = {Gu,  Jiaqi and Zhao,  Zheng and Feng,  Chenghao and Zhu,  Hanqing and Chen,  Ray T. and Pan,  David Z.},
  year = {2020},
  month = mar,
  pages = {1586–1589}
}

@article{Nahmias2020photonic,
  title = {Photonic Multiply-Accumulate Operations for Neural Networks},
  volume = {26},
  ISSN = {1558-4542},
  url = {http://dx.doi.org/10.1109/JSTQE.2019.2941485},
  DOI = {10.1109/jstqe.2019.2941485},
  number = {1},
  journal = {IEEE Journal of Selected Topics in Quantum Electronics},
  publisher = {Institute of Electrical and Electronics Engineers (IEEE)},
  author = {Nahmias,  Mitchell A. and de Lima,  Thomas Ferreira and Tait,  Alexander N. and Peng,  Hsuan-Tung and Shastri,  Bhavin J. and Prucnal,  Paul R.},
  year = {2020},
  month = jan,
  pages = {1–18}
}

@article{hamerly2019large,
  title = {Large-Scale Optical Neural Networks Based on Photoelectric Multiplication},
  volume = {9},
  ISSN = {2160-3308},
  url = {http://dx.doi.org/10.1103/PhysRevX.9.021032},
  DOI = {10.1103/physrevx.9.021032},
  number = {2},
  journal = {Physical Review X},
  publisher = {American Physical Society (APS)},
  author = {Hamerly,  Ryan and Bernstein,  Liane and Sludds,  Alexander and Soljačić,  Marin and Englund,  Dirk},
  year = {2019}
  }

@misc{murmann2022adc_survey,
  author = {Murmann, Boris},
  title = {{ADC Performance Survey 1997--2026}},
  year = {2026},
  howpublished = {\url{https://github.com/bmurmann/ADC-survey}},
  note = {Accessed: 2026-04-24}
}

@article{Williamson2020reprogrammable,
  title = {Reprogrammable Electro-Optic Nonlinear Activation Functions for Optical Neural Networks},
  volume = {26},
  ISSN = {1558-4542},
  url = {http://dx.doi.org/10.1109/JSTQE.2019.2930455},
  DOI = {10.1109/jstqe.2019.2930455},
  number = {1},
  journal = {IEEE Journal of Selected Topics in Quantum Electronics},
  publisher = {Institute of Electrical and Electronics Engineers (IEEE)},
  author = {Williamson,  Ian A. D. and Hughes,  Tyler W. and Minkov,  Momchil and Bartlett,  Ben and Pai,  Sunil and Fan,  Shanhui},
  year = {2020},
  month = jan,
  pages = {1–12}
}

@inproceedings{Bouchakour2025lowpower,
  title = {A Low Power Fully Analog MAC Operation Using Standard CMOS Technology for Neural Network Inference},
  url = {http://dx.doi.org/10.1109/NewCAS64648.2025.11106966},
  DOI = {10.1109/newcas64648.2025.11106966},
  booktitle = {2025 23rd IEEE Interregional NEWCAS Conference (NEWCAS)},
  publisher = {IEEE},
  author = {Bouchakour,  Mohamed and Bergeret,  Emmanuel and Sicard,  Gilles and Abdelouahab,  Kamel and Berry,  Fran\c{c}ois},
  year = {2025},
  month = june,
  pages = {10–14}
}

@misc{sze2017efficient,
      title={Efficient Processing of Deep Neural Networks: A Tutorial and Survey}, 
      author={Vivienne Sze and Yu-Hsin Chen and Tien-Ju Yang and Joel Emer},
      year={2017},
      eprint={1703.09039},
      archivePrefix={arXiv},
      primaryClass={cs.CV},
      url={https://arxiv.org/abs/1703.09039}, 
}

@misc{nvidia2024blackwell,
  author       = {{NVIDIA Corporation}},
  title        = {{NVIDIA Blackwell Architecture Technical Brief}},
  howpublished = {NVIDIA},
  year         = {2024},
  url          = {https://resources.nvidia.com/en-us-blackwell-architecture}
}

@InProceedings{ji2025rznas,
  title = 	 {{RZ}-{NAS}: Enhancing {LLM}-guided Neural Architecture Search via Reflective Zero-Cost Strategy},
  author =       {Ji, Zipeng and Zhu, Guanghui and Yuan, Chunfeng and Huang, Yihua},
  booktitle = 	 {Proceedings of the 42nd International Conference on Machine Learning},
  pages = 	 {27237--27254},
  year = 	 {2025},
  editor = 	 {Singh, Aarti and Fazel, Maryam and Hsu, Daniel and Lacoste-Julien, Simon and Berkenkamp, Felix and Maharaj, Tegan and Wagstaff, Kiri and Zhu, Jerry},
  volume = 	 {267},
  series = 	 {Proceedings of Machine Learning Research},
  month = 	 {13--19 Jul},
  publisher =    {PMLR},
  url = 	 {https://proceedings.mlr.press/v267/ji25a.html}
}

@inproceedings{nasir2024llmatic, 
   series={GECCO ’24},
   title={LLMatic: Neural Architecture Search Via Large Language Models And Quality Diversity Optimization},
   url={http://dx.doi.org/10.1145/3638529.3654017},
   DOI={10.1145/3638529.3654017},
   booktitle={Proceedings of the Genetic and Evolutionary Computation Conference},
   publisher={ACM},
   author={Nasir, Muhammad Umair and Earle, Sam and Togelius, Julian and James, Steven and Cleghorn, Christopher},
   year={2024},
   pages={1110–1118},
   collection={GECCO ’24} 
}

@article{Deb2002nsgaii,
  title = {A fast and elitist multiobjective genetic algorithm: NSGA-II},
  volume = {6},
  ISSN = {1089-778X},
  url = {http://dx.doi.org/10.1109/4235.996017},
  DOI = {10.1109/4235.996017},
  number = {2},
  journal = {IEEE Transactions on Evolutionary Computation},
  publisher = {Institute of Electrical and Electronics Engineers (IEEE)},
  author = {Deb,  K. and Pratap,  A. and Agarwal,  S. and Meyarivan,  T.},
  year = {2002},
  month = apr,
  pages = {182–197}
}

@misc{intel2021xeon8380,
  author       = {{Intel Corporation}},
  title        = {Intel Xeon Platinum 8380 Processor (60M Cache, 2.30 GHz) --- Product Specifications},
  howpublished = {Intel ARK (Product Information), SKU 212287},
  year         = {2021},
  url          = {https://www.intel.com/content/www/us/en/products/sku/212287/intel-xeon-platinum-8380-processor-60m-cache-2-30-ghz.html},
  note         = {Accessed: 2026-04-24}
}

@article{Krestinskaya2024imcnas,
  title = {Neural architecture search for in-memory computing-based deep learning accelerators},
  volume = {1},
  ISSN = {2948-1201},
  url = {http://dx.doi.org/10.1038/s44287-024-00052-7},
  DOI = {10.1038/s44287-024-00052-7},
  number = {6},
  journal = {Nature Reviews Electrical Engineering},
  publisher = {Springer Science and Business Media LLC},
  author = {Krestinskaya,  Olga and Fouda,  Mohammed E. and Benmeziane,  Hadjer and El Maghraoui,  Kaoutar and Sebastian,  Abu and Lu,  Wei D. and Lanza,  Mario and Li,  Hai and Kurdahi,  Fadi and Fahmy,  Suhaib A. and Eltawil,  Ahmed and Salama,  Khaled N.},
  year = {2024},
  month = May,
  pages = {374–390}
}

@InProceedings{wu2023quantumdarts,
  title = 	 {{Q}uantum{DARTS}: Differentiable Quantum Architecture Search for Variational Quantum Algorithms},
  author =       {Wu, Wenjie and Yan, Ge and Lu, Xudong and Pan, Kaisen and Yan, Junchi},
  booktitle = 	 {Proceedings of the 40th International Conference on Machine Learning},
  pages = 	 {37745--37764},
  year = 	 {2023},
  editor = 	 {Krause, Andreas and Brunskill, Emma and Cho, Kyunghyun and Engelhardt, Barbara and Sabato, Sivan and Scarlett, Jonathan},
  volume = 	 {202},
  series = 	 {Proceedings of Machine Learning Research},
  month = 	 {23--29 Jul},
  publisher =    {PMLR},
  url = 	 {https://proceedings.mlr.press/v202/wu23v.html}
}

@inproceedings{Shafiee2025luxnas,
  title = {LuxNAS: A Coherent Photonic Neural Network Powered by Neural Architecture Search},
  url = {http://dx.doi.org/10.1364/CLEO_AT.2025.JPS100_93},
  DOI = {10.1364/cleo_at.2025.jps100_93},
  booktitle = {CLEO 2025},
  publisher = {Optica Publishing Group},
  author = {Shafiee,  Amin and Sunny,  Febin and Pasricha,  Sudeep and Nikdast,  Mahdi},
  year = {2025}
}

@inproceedings{Martyniuk2024quantumnassurvey,
  title = {Quantum Architecture Search: A Survey},
  url = {http://dx.doi.org/10.1109/QCE60285.2024.00198},
  DOI = {10.1109/qce60285.2024.00198},
  booktitle = {2024 IEEE International Conference on Quantum Computing and Engineering (QCE)},
  publisher = {IEEE},
  author = {Martyniuk,  Darya and Jung,  Johannes and Paschke,  Adrian},
  year = {2024},
  month = Sept,
  pages = {1695–1706}
}

@InProceedings{na2022autosnn,
  title = 	 {{A}uto{SNN}: Towards Energy-Efficient Spiking Neural Networks},
  author =       {Na, Byunggook and Mok, Jisoo and Park, Seongsik and Lee, Dongjin and Choe, Hyeokjun and Yoon, Sungroh},
  booktitle = 	 {Proceedings of the 39th International Conference on Machine Learning},
  pages = 	 {16253--16269},
  year = 	 {2022},
  editor = 	 {Chaudhuri, Kamalika and Jegelka, Stefanie and Song, Le and Szepesvari, Csaba and Niu, Gang and Sabato, Sivan},
  volume = 	 {162},
  series = 	 {Proceedings of Machine Learning Research},
  month = 	 {17--23 Jul},
  publisher =    {PMLR},
  url = 	 {https://proceedings.mlr.press/v162/na22a.html}
}

@article{Yan2024efficientsnn,
  title = {Efficient spiking neural network design via neural architecture search},
  volume = {173},
  ISSN = {0893-6080},
  url = {http://dx.doi.org/10.1016/j.neunet.2024.106172},
  DOI = {10.1016/j.neunet.2024.106172},
  journal = {Neural Networks},
  publisher = {Elsevier BV},
  author = {Yan,  Jiaqi and Liu,  Qianhui and Zhang,  Malu and Feng,  Lang and Ma,  De and Li,  Haizhou and Pan,  Gang},
  year = {2024},
  month = May,
  pages = {106172}
}

@article{yu2025gptnas,
   title={GPT-NAS: Neural Architecture Search Meets Generative Pre-Trained Transformer Model},
   volume={8},
   ISSN={2097-406X},
   url={http://dx.doi.org/10.26599/BDMA.2024.9020036},
   DOI={10.26599/bdma.2024.9020036},
   number={1},
   journal={Big Data Mining and Analytics},
   publisher={Tsinghua University Press},
   author={Yu, Caiyang and Liu, Xianggen and Wang, Yifan and Liu, Yun and Feng, Wentao and Deng, Xiong and Tang, Chenwei and Lv, Jiancheng},
   year={2025},
   month=Feb, pages={45–64} 
}

@inproceedings{krishnakumar2022nasbenchsuitezero,
title={{NAS}-Bench-Suite-Zero: Accelerating Research on Zero Cost Proxies},
author={Arjun Krishnakumar and Colin White and Arber Zela and Renbo Tu and Mahmoud Safari and Frank Hutter},
booktitle={Thirty-sixth Conference on Neural Information Processing Systems Datasets and Benchmarks Track},
year={2022},
url={https://openreview.net/forum?id=yWhuIjIjH8k}
}

@article{Lecun1998lenet,
  title = {Gradient-based learning applied to document recognition},
  volume = {86},
  ISSN = {0018-9219},
  url = {http://dx.doi.org/10.1109/5.726791},
  DOI = {10.1109/5.726791},
  number = {11},
  journal = {Proceedings of the IEEE},
  publisher = {Institute of Electrical and Electronics Engineers (IEEE)},
  author = {Lecun,  Y. and Bottou,  L. and Bengio,  Y. and Haffner,  P.},
  year = {1998},
  pages = {2278–2324}
}

@misc{li2018hyperband,
      title={Hyperband: A Novel Bandit-Based Approach to Hyperparameter Optimization}, 
      author={Lisha Li and Kevin Jamieson and Giulia DeSalvo and Afshin Rostamizadeh and Ameet Talwalkar},
      year={2018},
      eprint={1603.06560},
      archivePrefix={arXiv},
      primaryClass={cs.LG},
      url={https://arxiv.org/abs/1603.06560}, 
}

@article{Su2024reliability,
  title = {Reliability analysis of optical neural networks with non-ideal signal transmission},
  volume = {87},
  ISSN = {1068-5200},
  url = {http://dx.doi.org/10.1016/j.yofte.2024.103928},
  DOI = {10.1016/j.yofte.2024.103928},
  journal = {Optical Fiber Technology},
  publisher = {Elsevier BV},
  author = {Su,  Ye and Fu,  Pengju and Ye,  Yichen and Chai,  Junxiong and Jiang,  Xiao and Yang,  Hongyu and Xie,  Yiyuan},
  year = {2024},
  month = Oct,
  pages = {103928}
}

\appendix

\section{Hardware Modeling Costs}

\subsection{Energy Efficiency}\label{app:energy_efficiency}

For GPU inference, we estimate per-inference energy from an NVIDIA B200 Blackwell GPU operating at ideal FP16 tensor core utilization. Assuming a thermal design power of 1000~W and a peak throughput of $2.25 \times 10^{15}$ FLOPS, each FP16 multiply-accumulate dissipates approximately $\tfrac{2 \times 1000~\mathrm{W}}{2.25 \times 10^{15}~\mathrm{FLOPS}} \approx 0.89$~pJ/MAC \cite{nvidia2024blackwell}. The same calculation under FP8 doubles the throughput to $4.5 \times 10^{15}$ FLOPS, yielding a more aggressive $\sim 0.44$~pJ/MAC; we use the FP16 figure throughout as it more closely matches the precision regime in our experiments. Element-wise nonlinear activations are estimated at 3~pJ/element, consistent with prior energy accounting for digital accelerators \cite{sze2017efficient}.

For CPU inference, we estimate per-inference energy from an Intel Xeon Platinum 8380 running single-threaded FP32 inference. Peak FP32 throughput is derived as $40~\text{cores} \times 2~\text{FMA units} \times 16~\text{lanes} \times 2~\text{ops} \times 2.3~\mathrm{GHz} \approx 5.89~\mathrm{TFLOPS}$, which combined with a 270~W TDP yields an FP32 MAC cost of $\tfrac{2 \times 270~\mathrm{W}}{5.89 \times 10^{12}~\mathrm{FLOPS}} \approx 91.7$~pJ/MAC \cite{intel2021xeon8380}. As with the GPU, we charge 3~pJ per element for ReLU and other element-wise nonlinearities \cite{sze2017efficient}. The roughly two-orders-of-magnitude gap between CPU and GPU MAC energy reflects the absence of dedicated tensor units and the lower arithmetic intensity of general-purpose CPU pipelines.

For the MZI-based ONN, we leverage a simple estimator that analytically computes per-inference energy based on the network's layer structure. Linear and convolutional operations executed on the MZI mesh are charged at 20~fJ/MAC, an estimate consistent with reported energy figures for silicon-photonic tensor cores \cite{Nahmias2020photonic, hamerly2019large, Hamerly2022scalable}. Because MZI meshes are inherently linear, every signal entering or leaving the optical domain incurs a data conversion cost: 2~pJ/element for input digital-to-analog conversion \cite{Bouchakour2025lowpower} and 4~pJ/element for output analog-to-digital conversion \cite{murmann2022adc_survey}. Element-wise nonlinearities are performed in the electronic domain at 10~pJ/element \cite{Williamson2020reprogrammable}, a substantially higher per-element cost than on digital hardware that reflects the overhead of opto-electronic conversion at each nonlinearity. The dominant per-MAC advantage of the optical regime is therefore partially offset by these fixed conversion costs, which become amortized for deeper or wider networks.

\subsection{Non-ideality Model}\label{app:non_idealities}

For tested hardware, we consider multiple relevant factors, including quantization, hardware constraints, and hardware non-idealities in an attempt to closely simulate expected hardware performance. In the case of CPU and GPU, we note that there are no non-idealities that need to be considered.

For optical hardware, we simulate the performance of MZI meshes for deep learning as seen in Shen et al. \cite{Shen2017nanophotonics}, leveraging pytorch-onn \cite{jiaqigu2021L2ight} to implement MZI-based linear and convolutional layers. To accurately represent bit precision in photonic MZI meshes, we simulate 8-bit precision for MZI phase values and input/output signals \cite{Shen2017nanophotonics, Demirkiran2023electrophotonic}, training with quantization-aware training (QAT). 

We additionally impose architectural constraints pertinent to MZI meshes. These include block-diagonal decomposition, decomposing weight matrices into $4 \times 4$ sub-matrices implemented as a Clements-topology MZI mesh \cite{Vadlamani2023transferable}. We also constrain the width of linear layers to $\leq 512$ (a soft physical constraint) based on the maximally demonstrated MZI meshes \cite{Hamerly2022scalable}, along with heavily penalizing residual connections because these require additional ADC/DAC conversions to allow signal summation. Due to the lack of nonlinear activations in optical computing, we require opto-electronic conversions that induce higher energy costs. Finally, we include ADC/DAC costs at networks input/output layers, resulting in negligible energy overhead for deeper networks. 

For hardware non-idealities, we capture four relevant physical noise sources that typically arise in silicon-photonic chips (and that are provided through pytorch-onn's open-source framework). These include phase error, crosstalk, gamma noise, and quantization noise (the last of which is handled by our existing QAT scheme). Phase error represents Gaussian perturbations on MZI phase shifter values arising from fabrication imprecision, where $\sigma_\phi \in [0.01, 0.05]$ is a reasonable estimate for MZI ONNs \cite{Shen2017nanophotonics, Pai2023backprop}. We assume thermal crosstalk of $\varepsilon \in [0.01, 0.15]$ (modeled as nearest-neighbor weight-space leakage), which is consistent with estimates for silicon-photonic MZI meshes \cite{Hamerly2022scalable, Milanizadeh2019canceling}. Finally, for gamma noise, we represent it as $\sigma_\gamma \in [0.001, 0.005]$, which captures a multiplicative per MZI voltage-to-phase conversion error \cite{Gu2020roq}. These non-idealities are injected by perturbing trained weight tensors in weight-space. These outputs are then evaluated on the validation set to produce non-ideal accuracy metrics, which are then fed into the LLM's knowledge base for future generations.



\section{Energy Estimates and Validation}

\begin{figure}[t]
  \centering
\includegraphics[width=\columnwidth]{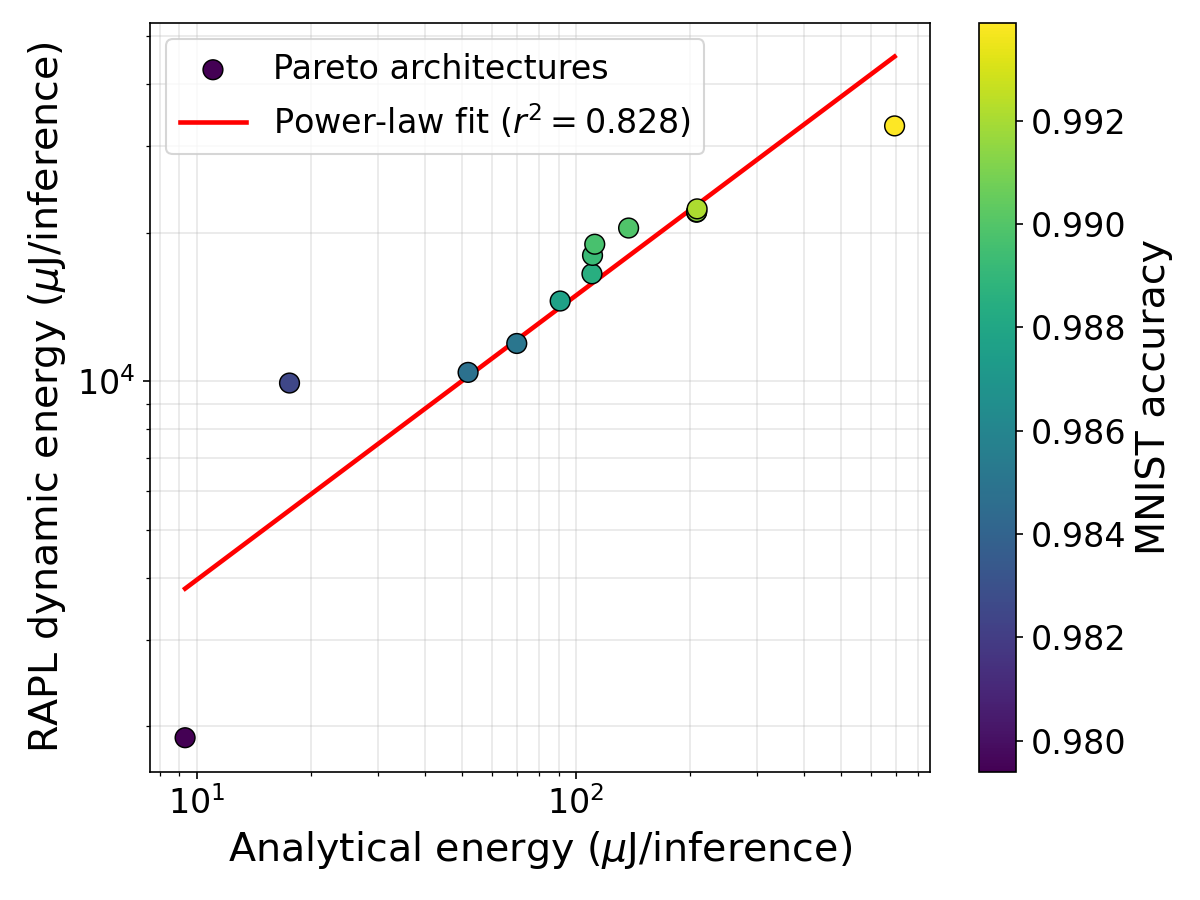}
    \caption{Log-log correlation between analytical energy estimates and real RAPL energy validation on an AWS \texttt{c6i.metal} instance with Xeon Platinum 8375C (Ice Lake-SP), 2.90\,GHz, single core. Only models along the Pareto frontier were tested, giving a log-space correlation of $r^2 = 0.828$. Averaged over 50{,}000 forward passes. }
  \label{fig:rapl-correlation}
\end{figure}


All hardware backends in this work currently compute per-inference energy from a closed-form analytical model described in Appendix \ref{app:energy_efficiency}. These estimations are essential to the NSGA-II loop, as the evaluated candidates are sorted by both energy efficiency as well as accuracy. Although our numbers are pulled directly from product specifications, the question of how accurate these estimates are remains. To test this, we validate energy results by re-running every architecture along the CPU Pareto frontier on bare-metal Xeon silicon AWS \texttt{c6i.metal} instances, utilizing Intel's RAPL \texttt{MSR\_PKG\_ENERGY\_STATUS} counter averaged over $N=50{,}000$ batch-size-1 passes. Because AWS does not expose B200 instances at the time of writing, only CPU energy estimates are included. 

Figure~\ref{fig:rapl-correlation} and Table~\ref{tab:rapl-validation} compare the results between the closed-form analytical model and the real energy results. Although the analytical estimates under-estimate by 2 orders of magnitude on average, we note that the ranking fidelity is high: when using a least-squares fit, we obtain an $r^2$ of 0.828. 

This offset is expected because the analytical model assumes best-case performance, and thus serves as a theoretical maximum bound. Because PyTorch inference induces additional computational overhead, the RAPL energy costs tend to be much higher. However, we note that every hardware backend tested in UH-NAS utilizes the same theoretical maximum convention, and thus cross-platform comparisons are still valid. It is worth noting then that architecture rankings by Pareto frontiers are best served as theoretical maximum comparisons, not as deployment-time comparisons. 

\begin{table*}[t]
  \centering
  \caption{RAPL energy validation of CPU Pareto architectures on Intel Xeon Platinum 8375C (Ice Lake, single-threaded, single-core). Analytical energy uses $E_{\mathrm{MAC}} = 91.7$ pJ/MAC~\cite{intel2021xeon8380}. RAPL dynamic energy subtracts measured idle power (59.6 W) over the inference interval. Each entry averages 50{,}000 forward passes on batch size 1 MNIST images.}
  \label{tab:rapl-validation}
  \small
  \begin{tabular}{@{}l l c c c c c@{}}
  \toprule
  Acc.\ (\%) & Architecture & Params & Analytical ($\mu$J) & RAPL Dyn.\ ($\mu$J) & Ratio & Time
  ($\mu$s) \\
  \midrule
  97.94 & 2-layer MLP                         & 101K & 9.3   & 1{,}898  & 204$\times$ & 36  \\
  98.25 & 1-conv + 1-res, stride-2 head       & 6K   & 17.6  & 9{,}925  & 565$\times$ & 192 \\
  98.48 & 4-layer MLP (wide)                  & 569K & 52.0  & 10{,}431 & 201$\times$ & 191 \\
  98.51 & 1-conv + 1-res, FC head             & 6K   & 69.8  & 11{,}936 & 171$\times$ & 229 \\
  98.77 & 1-conv + 1-res + 1-conv, FC head    & 14K  & 90.9  & 14{,}551 & 160$\times$ & 280 \\
  98.85 & 1-conv + 2-res, minimal head        & 13K  & 110.2 & 16{,}519 & 150$\times$ & 318 \\
  98.92 & 1-conv + 2-res, 1-FC head           & 18K  & 110.6 & 18{,}001 & 163$\times$ & 346 \\
  98.97 & 1-conv + 2-res, 2-FC head           & 34K  & 112.1 & 18{,}965 & 169$\times$ & 364 \\
  98.99 & 1-conv + 2-res, 3-FC head (wide)    & 59K  & 137.7 & 20{,}454 & 149$\times$ & 392 \\
  99.13 & 2-conv + 2-res, narrow FC head      & 27K  & 208.0 & 22{,}019 & 106$\times$ & 421 \\
  99.14 & 2-conv + 2-res, medium FC head      & 30K  & 208.2 & 22{,}066 & 106$\times$ & 421 \\
  99.21 & 2-conv + 2-res, wide FC head        & 36K  & 208.7 & 22{,}365 & 107$\times$ & 427 \\
  99.39 & 2-conv + 2-res (deep), 2-FC head    & 91K  & 692.3 & 32{,}933 & 48$\times$  & 620 \\
  \bottomrule
  \end{tabular}
  \vspace{0.5em}

  \raggedright\footnotesize
  \textit{Platform:} AWS \texttt{c6i.metal}, Xeon Platinum 8375C (Ice Lake-SP), 2.90\,GHz, single core. \\
  \textit{Method:} RAPL energy status read before/after $N{=}50{,}000$ inferences; idle power measured over a 2 second interval and subtracted. \\
  \end{table*}

\section{System Prompt Analysis}\label{app:system_prompt}

UH-NAS leverages hardware-specific system prompts to condition the LLM on physical constraints, non-idealities, and energy-aware design principles during neural architecture search. Representative prompts synthesized by GPT-4.1-nano, GPT-4.1, and Gemini 3.1 Flash-Lite are shown in Figures~\ref{fig:nano-prompt-example}, \ref{fig:gpt41-prompt-example}, and \ref{fig:gemini-prompt-example} for the optical MZI task. These system prompts are synthesized once per run after initial evaluation of seed architectures and include both physically-motivated heuristics and design tips. 

Stronger models such as GPT-4.1 (Figure~\ref{fig:gpt41-prompt-example}) and Gemini 3.1 Flash-Lite (Figure~\ref{fig:gemini-prompt-example}) both produce prompts that are factually grounded in hardware energy hierarchy, effectively guiding the end-to-end NAS. Although GPT-4.1 is markedly more verbose than Gemini 3.1 Flash-Lite, they reach the same conclusions: optical MACs are essentially free relative to the electronic ReLU activations, and thus good architectures should maximize optical MACs to reduce the number of electronic boundary crossings.

By comparison, GPT-4.1-nano (Figure~\ref{fig:nano-prompt-example}) inverts the signal, causing the initial seed architectures to frequently be the best observed architectures throughout the entire NAS (as seen in Figure~\ref{fig:accuracy_generation}). This is primarily caused by an erroneous statement in the system prompt: ``Prioritize minimizing
optical MAC counts and electronic activation energy''. This is propagated by the design tips (\#3 ``minimize the number of MZI components by avoiding overly
wide layers''; \#7 ``reduce the total number of parameters and layer
widths''), which in parallel encourage the minimization of both linear MACs as well as electronic ReLUs. This results in significantly narrower and shallower networks that are incapable of learning relevant signals in the worst-case non-ideal MNIST task, leading to degenerate Pareto frontiers.

\begin{figure*}[t]
  \begin{tcolorbox}[
    title={\textbf{Synthesized System Prompt via GPT-4.1-nano}},
    colback=gray!5,
    colframe=gray!60,
    fonttitle=\small\bfseries
  ]
  \small
  \textbf{System Prompt:}\\[4pt]
  \textit{You are a hardware-aware neural architecture designer focusing on optimizing neural network architectures for the torchonn optical backend characterized by phase noise, crosstalk, and quantization constraints. Your goal is to develop architectures that balance non-idealities and energy efficiency, leveraging published principles in photonic and electronic hardware design. \underline{Prioritize minimizing optical MAC counts} and electronic activation energy, while mitigating the impact of crosstalk and phase noise on model accuracy. Remember that ADC/DAC costs are only incurred at the network boundaries, so layer-wise domain crossings should be minimized and arranged to maximize optical processing. Use your understanding of physical mechanisms---such as crosstalk coupling within miniblocks, phase noise sensitivity, and quantization effects---to guide architecture choices, ensuring dimension and layer constraints are satisfied. Your outputs should be architecture strings optimized for the given hardware regime, reflecting these principles.}
  \vspace{6pt}
  \textbf{Design Tips:}
  \begin{enumerate}[leftmargin=*, itemsep=2pt, parsep=0pt]
    \item Design linear blocks to be as shallow as possible and avoid unnecessary ReLU activations within the optical domain to reduce electronic energy and phase noise sensitivity.
    \item Prioritize optical linear and convolutional operations over electronic ones, as they consume significantly less energy (${\sim}20$~fJ/MAC), and ensure they are arranged after the transition point (Flatten or AdaptPool).
    \item Minimize the number of MZI components by avoiding overly wide layers, since MZI count scales quadratically; choose layer widths that balance expressivity with hardware complexity.
    \item Reduce crosstalk impact by limiting miniblock output coupling---prefer architectures with fewer adjacent outputs within each miniblock or include dropout to improve crosstalk tolerance.
    \item Leverage convolutional layers before the transition point to extract spatial features efficiently, as they do not incur ADC/DAC penalties, and keep the depth of the network shallow to mitigate phase noise accumulation.
    \item Incorporate BN layers within the optical domain to enhance phase-noise robustness without incurring electronic activation costs, but avoid excessive BN or Dropout that could increase complexity and crosstalk sensitivity.
    \item Reduce the total number of parameters and layer widths while maintaining sufficient representational capacity to accommodate the non-idealities, thus controlling crosstalk and phase noise effects.
    \item Use $1{\times}1$ convolutions selectively for channel mixing to keep the model compact and reduce the number of optical elements, especially in deeper layers.
    \item Design the network with a single transition point followed by MZI-based linear blocks to limit electronic boundary crossings, thereby conserving energy and reducing crosstalk pathways.
  \end{enumerate}
  \end{tcolorbox}
  \caption{Example synthesized system prompt from the meta-learning phase (GPT-4.1-nano). The underlined sentence contains a factual error: namely that optical MACs at${\sim}20$~fJ are cheap and should be maximized, not minimized due to their low computational cost. }
  \label{fig:nano-prompt-example}
\end{figure*}

\begin{figure*}[t]
\begin{tcolorbox}[
  title={\textbf{Synthesized System Prompt via GPT-4.1}},
  colback=gray!5,
  colframe=gray!60,
  fonttitle=\small\bfseries
]
\small
\textbf{System Prompt:}\\[4pt]
\textit{You are a hardware-aware neural architecture designer targeting integrated silicon photonic mesh accelerators, implemented via the torchonn backend. Your objective is to search for neural architectures that maximize task accuracy under realistic non-idealities of the photonic hardware, while minimizing total energy consumption. You must account for physical noise sources---such as phase noise, gamma noise, and crosstalk between adjacent outputs within each miniblock---that degrade inference accuracy compared to idealized models. Your designs will be realized using MZI-based linear and convolutional blocks, with batch normalization and dropout as key tools to mitigate the impact of hardware noise.}

\textit{In this regime, noise and signal integrity are significant: phase noise (std=0.05) and crosstalk (factor=0.15) represent moderate-to-severe non-idealities, as seen in experimental silicon photonic mesh reports. Digital resolution is limited (8-bit weights/inputs, miniblock=4), further constraining representational fidelity. The dominant energy cost arises from electronic ReLU activations, not from photonic MACs. Residual connections are unavailable due to the prohibitive cost and complexity of frequent ADC/DAC crossings. Therefore, you must seek architectures that are robust to analog noise, judicious in ReLU/BN placement, and efficient in both parameter count and energy.}

\textit{Follow design strategies from the hardware-efficient ML and silicon photonics literature (e.g., work on MZI mesh robustness, photonic neural networks, analog noise mitigation via normalization and dropout, and energy-aware NN design). Your choices should be grounded in physical principles---such as how crosstalk, quantization, and nonlinearity placement affect accuracy and efficiency on analog photonic substrates.}

\textit{You are not to prescribe or prefer specific dimensions or depths, but to articulate physically-motivated heuristics for constructing architectures that maximize robustness and efficiency within the constraints of mesh-based photonic inference.}

\vspace{6pt}
\textbf{Design Tips:}
\begin{enumerate}[leftmargin=*, itemsep=2pt, parsep=0pt]
  \item Prioritize batch normalization (BN) immediately after optical MZI layers to mitigate phase noise and offset drift, as shown in photonic NN literature.
  \item Introduce dropout after MZI layers to improve tolerance to crosstalk and random phase errors---dropout acts as a regularizer against structured analog noise.
  \item Minimize the number of electronic ReLU activations, since their per-element energy cost far exceeds that of optical MACs; use them only where nonlinearity is crucial for expressivity.
  \item Exploit the efficiency of Conv2d blocks for early spatial feature extraction, as they are native to the hardware and do not incur extra ADC/DAC energy penalty.
  \item Be aware that crosstalk in MZI meshes couples adjacent outputs within each miniblock---avoid excessive width increases that exacerbate cumulative crosstalk.
  \item Constrain layer widths and output channels to multiples of 8 (except the classifier head) to align with hardware-friendly quantization and mesh partitioning.
  \item Use only a single transition point (Flatten or AdaptPool) between Conv and MZI blocks, respecting the mesh's lack of native support for skip/residual connections.
  \item Favor architectures that separate spatial processing (Conv2d) and dense classification (MZI Linear), leveraging each domain's strengths and minimizing unnecessary domain crossings.
  \item Remember that parameter count scales quadratically with MZI layer width---balance model capacity with physical mesh resource limits.
  \item Consider the compound effect of quantization noise (8b weights/inputs) and analog noise sources when stacking multiple linear/MZI layers; normalization and dropout help preserve signal integrity.
\end{enumerate}
\end{tcolorbox}
\caption{Example synthesized system prompt from the meta-learning phase (GPT-4.1). Unlike the GPT-4.1-nano prompt (Figure~\ref{fig:nano-prompt-example}), this prompt correctly identifies electronic ReLU as the dominant energy cost, provides exact non-ideality parameters (phase noise std=0.05, crosstalk factor=0.15, 8-bit quantization, miniblock=4), and articulates physically-grounded heuristics without factual errors.}
\label{fig:gpt41-prompt-example}
\end{figure*}

\begin{figure*}[t]
\begin{tcolorbox}[
  title={\textbf{Synthesized System Prompt via Gemini 3.1 Flash-Lite}},
  colback=gray!5,
  colframe=gray!60,
  fonttitle=\small\bfseries
]
\small
\textbf{System Prompt:}\\[4pt]
\textit{You are an expert Photonic Integrated Circuit (PIC) architect specializing in MZI-mesh neural accelerators. Your objective is to design high-accuracy, noise-resilient architectures that navigate the severe hardware constraints of the torchonn backend. You must treat the network as a monolithic optical signal path where electronic bottlenecks---specifically the 10~pJ/element ReLU activations---are treated as primary energy drivers rather than cost-free operations.}

\textit{Your design philosophy is grounded in the reality of photonic hardware: phase noise and crosstalk are cumulative, and the $O(N^2)$ MZI scaling makes excessive layer width physically prohibitive. You must balance the expressive power of non-linear activations against their energy cost and noise-magnification properties. Because residual connections are unavailable, you must prioritize stable, shallow-to-medium depth architectures that maintain signal integrity through hardware-aware regularization and noise-tolerant layer choices.}

\vspace{6pt}
\textbf{Design Tips:}
\begin{enumerate}[leftmargin=*, itemsep=2pt, parsep=0pt]
  \item Prioritize BN (Batch Normalization) in every MZI layer to normalize signal variance, which is critical for mitigating the impact of phase noise (std=0.05) on weight precision.
  \item Minimize the use of ReLU activations; since they cost ${\sim}10$~pJ/element vs ${\sim}20$~fJ/MAC for optical operations, use them only where non-linearity is strictly required for convergence.
  \item Exploit the miniblock ($k{=}4$) structure by grouping highly correlated features together, as crosstalk is localized within these miniblocks.
  \item Use MZILinearDrop to combat the 0.15 crosstalk factor; dropout acts as a structural regularizer that forces the network to learn redundant representations less sensitive to local hardware fluctuations.
  \item Favor ConvK1BNReLU for channel mixing over massive linear layers to keep the MZI count ($O(N^2)$) manageable and reduce the accumulation of phase noise.
  \item Balance the depth of the Conv front-end against the MZI back-end; deeper architectures increase the total MZI count, thereby multiplying the cumulative phase noise impact.
  \item When non-ideal accuracy drops significantly, prefer widening the layer slightly over increasing depth, as depth leads to faster signal degradation due to the lack of skip connections.
  \item Prefer AdaptPool over Flatten to reduce the feature dimension before entering the MZI-mesh head, drastically lowering the energy cost and MZI footprint of the final Linear layers.
\end{enumerate}
\end{tcolorbox}
\caption{Example synthesized system prompt from the meta-learning phase (Gemini 3.1 Flash-Lite). This prompt is notably more concise than the GPT-4.1 variant (Figure~\ref{fig:gpt41-prompt-example}) while achieving comparable non-ideal accuracy (${\sim}97\%$). It embeds quantitative hardware parameters directly into the design tips rather than the system prompt preamble, and explicitly frames the width-vs-depth tradeoff in terms of signal degradation.}
\label{fig:gemini-prompt-example}
\end{figure*}



\end{document}